\documentclass[11pt,a4paper]{article}
\usepackage{authblk}
\usepackage[hyperref]{acl2021}
\usepackage{times}
\usepackage{latexsym}
\usepackage{graphicx}
\usepackage{multirow}
\usepackage{lipsum,afterpage,refcount}

\usepackage{microtype}
\usepackage[linewidth=1pt]{mdframed}

\aclfinalcopy 

\title{Recent Advances in Natural Language Processing via Large Pre-Trained Language Models: A Survey}

\author[1]{Bonan Min*}
\author[2]{Hayley Ross*}
\author[3]{Elior Sulem*}
\author[4]{Amir Pouran Ben Veyseh*}
\author[4]{\\Thien Huu Nguyen}
\author[5]{Oscar Sainz}
\author[5]{Eneko Agirre}
\author[1]{Ilana Heinz}
\author[3]{Dan Roth}
\affil[1]{Raytheon BBN Technologies}
\affil[ ]{\texttt{\{bonan.min, ilana.Heintz\}@raytheon.com}}
\affil[2]{Harvard University}
\affil[ ]{\texttt{hayleyross@g.harvard.edu}}
\affil[3]{University of Pennsylvania}
\affil[ ]{\texttt{\{eliors, danroth\}@seas.upenn.edu}}
\affil[4]{University of Oregon}
\affil[ ]{\texttt {\{apouran, thien\}@cs.uoregon.edu}} 
\affil[5]{University of the Basque Country (UPV/EHU)}
\affil[ ]{\texttt {\{oscar.sainz, e.agirre\}@ehu.eus}} 
\affil[ ]{} 
\affil[ ]{* indicate equal contribution}

\date{}

\begin{document}
\maketitle
\begin{abstract}
Large, pre-trained transformer-based language models such as BERT have drastically changed the Natural Language Processing (NLP) field. We present a survey of recent work that uses these large language models to solve NLP tasks via pre-training then fine-tuning, prompting, or text generation approaches. We also present approaches that use pre-trained language models to generate data for training augmentation or other purposes. We conclude with discussions on limitations and suggested directions for future research. 
\end{abstract}

\section{Introduction}

In recent years, large pre-trained transformer-based language models (PLMs), such as the BERT \citep{devlin-etal-2019-bert} and GPT \citep{radford2018improving} families of models, have taken Natural Language Processing (NLP) by storm, achieving state-of-the-art performance on many tasks. 

These large PLMs have fueled a paradigm shift in NLP. Take a classification task $p(y|x)$ (classifying textual input $x$ into a label $y$) as an example: traditional statistical NLP approaches often design hand-crafted features to represent $x$, and then apply a machine learning model (e.g. SVM \citep{cortes1995support}, logistic regression) to learn the classification function. Deep learning models learn the latent feature representation via a deep neural network \cite{lecun2015deep} in addition to the classification function. Note that the latent  representation needs to be learned afresh for each new NLP task, and that, in many cases, the size of the training data limits the quality of the latent feature representation. Given that the nuances of language are common to all NLP tasks, one could posit that we could learn a generic latent feature representations from some generic task once, and then share it across all NLP tasks. Language modeling, where the model needs to learn how to predict the next word given previous words, is  such a generic task with abundant naturally occurring text to pre-train such a model (hence the name pre-trained language models).  In fact, the latest, ongoing paradigm shift begins when PLMs are introduced: for numerous NLP tasks, researchers now leverage existing PLMs via {\it fine-tuning} for the task of interest, {\it prompting} the PLMs to perform the desired task, or reformulating the task as a {\it text generation} problem with application of PLMs to solve it accordingly. Advances in these three PLM-based paradigms have continuously established new state-of-the-art performances. 

This paper surveys recent works that leverage PLMs for NLP. We organize these works into the following three paradigms: 

\begin{itemize}
    \item Pre-train then fine-tune (\S ~\ref{sec:paradigm1}): perform general-purpose pre-training with a large unlabeled corpus, and then perform a small amount of task-specific fine-tuning for the task of interest.
    \item Prompt-based learning (\S ~\ref{sec:paradigm2}): prompt a PLM such that solving an NLP task is reduced to a task similar to the PLM's pre-training task (e.g. predicting a missing word), or a simpler proxy task (e.g. textual entailment). Prompting can usually more effectively leverage the knowledge encoded in the PLMs, leading to few-shot approaches. 
    \item NLP as text generation (\S ~\ref{sec:paradigm3}): Reformulate NLP tasks as text generation, to fully leverage knowledge encoded in a generative language model such as GPT-2~\cite{radford2019language} and T5~\cite{raffel2020exploring}. 
\end{itemize}

Generative PLMs can be also used for text generation tasks. We refer readers to the excellent surveys on text generation such as ~\citet{li2021pretrained} and ~\citet{yu2021survey}. This paper, unless otherwise specified, focuses on tasks that are not generative in nature (e.g. classification, sequence labeling  and structure prediction) that still cover a broad range of NLP tasks including syntactic or semantic parsing of text, Information Extraction (IE), Question Answering (QA), Textual Entailment (TE), sentiment analysis, and so on. 

In addition to the three paradigms, there is another, complementary method: to indirectly use any of the PLM paradigms above to improve results of target NLP tasks:

\begin{itemize}
    \item Data generation (\S ~\ref{sec:data_gen}):  run PLMs to automatically generate data for NLP tasks. The generated data can be silver labeled data, where typically the generative PLM is fine-tuned for the task, or some auxiliary data, such as counterexamples, clarifications, contexts, or other. In the first case, the silver labeled data can be added to existing labeled data. In the second case, the auxiliary data supports the target task in some way.
\end{itemize} 

The paper is organized as follows: Section \ref{sec:paradigm1} provides background on the PLMs and describes the first paradigm, {\it pre-train then fine-tune}. Section~\ref{sec:paradigm2} discusses the second paradigm, {\it prompt-based learning}. Section~\ref{sec:paradigm3} summarizes works in the third paradigm, {\it NLP as text generation}. In Section~\ref{sec:data_gen}, we describe approaches that generate data via PLMs for a broad range of NLP tasks. We discuss limitations and provide directions for future research in Section~\ref{sec:discussion} and conclude in Section~\ref{sec:conclusion}.

\section{Paradigm 1: Pre-Train then Fine-Tune} \label{sec:paradigm1}


\begin{figure*}
  \centering 
  \includegraphics[scale=0.46]{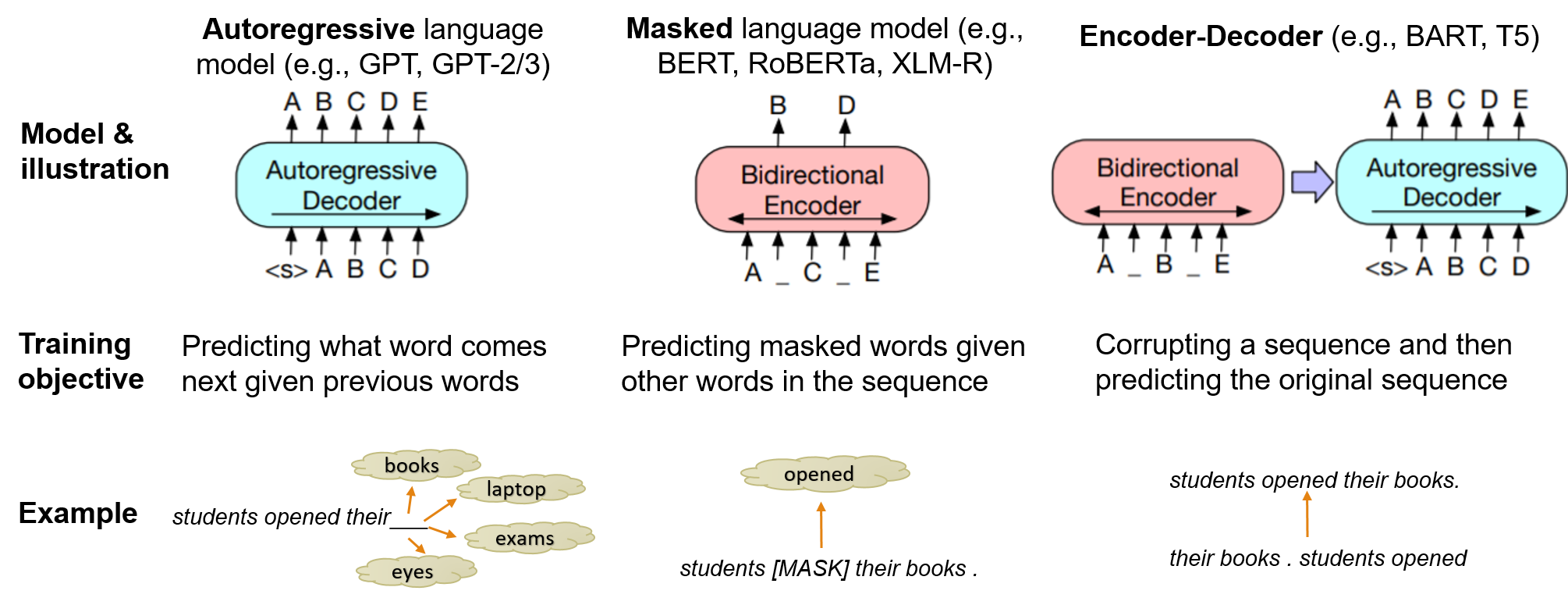}
  \caption{Three types of pre-trained language models. Model architecture illustrations are from~\citet{lewis-etal-2020-bart}. For the encoder-decoder model, the corruption strategy of document rotation is shown. Alternatives include sentence permutation, text infilling, token deletion/masking, etc.}
  \label{fig:plm_arch}
\end{figure*}




While work in traditional statistical NLP focused on training task-specific models on labeled datasets, this paradigm shifts to training one large model on a shared, ``fundamental'' pre-training task and then adapting (``fine-tuning'') it to a variety of tasks in a second step. The pre-training task is almost invariably a type of language modeling task\footnote{The exact formulation varies from the classic unidirectional language modeling (next word prediction) to cloze-style fill-in-the-blank, uncorrupting spans, and other variants (see Section \ref{sec:paradigm1-pretraining}).} that can leverage a massive quantity of unlabelled data to learn representations that benefit a range of NLP tasks~\cite{rogers-etal-2020-primer}.

In this section, we first provide a primer on pre-trained large language models (PLMs), then describe approaches that use frozen or fine-tuned PLMs for NLP tasks. 


\subsection{The Beginnings of the Paradigm Shift}

While pre-training in machine learning and, in particular, computer vision has been studied since at least 2010 \citep{erhan2010why,yosinski2014how,huh2016what},
the technique did not gain traction in NLP until later in the decade, with the publication of \citet{vaswani2017attention}. The delay in uptake is partly due to the later arrival of deep neural models to NLP compared to computer vision, partly due to the difficulty of choosing a self-supervised task\footnote{In self-supervised learning, the ground truth (e.g. the missing word) comes from the unlabeled text itself. This allows the pre-training to scale up with the near-infinite amount of text available on the web.} suitable for pre-training, and above all, due to the need for drastically larger model sizes and corpora in order to be effective for NLP tasks. We explore these aspects further in the discussion below.  

The idea of pre-training on a language modeling task is quite old. \citet{collobert2008unified} first suggested pre-training a model on a number of tasks to \emph{learn} features instead of hand-crafting them  (the predominant approach at the time). Their version of language model pre-training, however, differed significantly from the methods we see today. They used language modeling as only one of many tasks in a multitask learning setting, along with  other supervised tasks such as part-of-speech (POS) tagging, named entity recognition (NER) and semantic role labeling (SRL). 
\citeauthor{collobert2008unified} proposed sharing the weights of their deepest convolutional layer -- the word embeddings learned by the model -- between the multiple training tasks and fine-tuning the weights of the two remaining two feed-forward layers for each individual task.

Pre-training and fine-tuning did not gain popularity in NLP until the advent of ELMo \citep{peters2018deep} and ULMFiT \citep{howard2018universal}. Both models are based on Long Short-Term Memory architecture (LSTMs)~\cite{hochreiter1997long}, but differ in significant ways. ULMFiT pre-trains a three-layer LSTM 
on a standard language modeling objective, predicting the next token in a sequence.  ELMo uses layers of bidirectional LSTMs that combine two language model tasks in forward and backward directions to capture context from both sides. Both proposed fine-tuning the language model layer by layer for downstream application. Both studies also suggested adding additional classifier layers on top of the language model, which were fine-tuned alongside the language model layers.
These changes, combined with the substantially larger model size and pre-training corpus size compared to previous models, allowed the pre-training then fine-tuning paradigm to succeed. Both ELMo and ULMFiT showed competitive or improved performance compared to the then-state-of-the-art for a number of tasks, demonstrating the value of language model pre-training on a large scale.

The pace of this paradigm shift picked up dramatically in late 2018 when \citet{vaswani2017attention} introduced the Transformer architecture that can be used for language model pre-training. The Transformer's multi-head self-attention mechanism allows every word to attend to all previous words or every word except the target, allowing the model to efficiently capture long-range dependencies without the expensive recurrent computation in LSTMs. Multiple layers of multi-head self-attention allow for increasingly more expressive representations, useful for a range of NLP problems. As a result, nearly all popular language models, including GPT, BERT, BART \citep{lewis-etal-2020-bart} and T5 \citep{raffel2020exploring}, are now based on the Transformer architecture. They also differ in a number of important ways, which we discuss in the following sections. For more details about the Transformer architecture, we refer the reader to the original paper or to the excellent tutorials available\footnote{\url{http://nlp.seas.harvard.edu/2018/04/03/attention.html}}$^,$\footnote{\url{http://jalammar.github.io/illustrated-transformer/}}.


\subsection{Modern Pre-Trained Language Models}

There are three classes of pre-trained language models: autoregressive language models (e.g. GPT), masked language models (e.g. BERT), and encoder-decoder models (e.g. BART, T5). Figure~\ref{fig:plm_arch} shows the difference in model architecture and training objectives with an example training input for each. 


\begin{table*}[]
    \centering
\scalebox{0.81}{
\begin{tabular}{|l|c|c|c|}
  \hline
\textbf{Model} & \textbf{Pre-Training Sources} & \textbf{Size of Pre-Training Corpus} & \textbf{\# Model parameters} \\
\hline
\multicolumn{4}{|c|}{(1) \textbf{English Monolingual Models}} \\
\hline
\textsc{BERT(Base)}\cite{devlin-etal-2019-bert} & Wiki, books & 3.3B tokens (13GB data) & 110M \\
\hline
\textsc{BERT(Large)\cite{devlin-etal-2019-bert}} & Wiki, books & 3.3B tokens (13GB data) & 340M \\
\hline
\textsc{RoBERTa}\cite{liu2019roberta} & Wiki, books, web crawl & 161GB data & 340M \\ 
\hline
\textsc{XLNet}~\citep{yang2019xlnet} & Wiki, books, web crawl & 142GB data & 340M \\
\hline
\textsc{GPT}\citep{radford2018improving} & Web crawl & 800M tokens & 117M \\
\hline
\textsc{GPT-2}\cite{radford2019language} & Web crawl & 8M documents (40GB data) & 1.5B \\
\hline
\textsc{GPT-3}\cite{NEURIPS2020_1457c0d6} & Wiki, books, web crawl & 500B tokens & 175B \\ 
\hline
\textsc{BART}~\cite{lewis-etal-2020-bart} & Wiki, books & 3.3B tokens & $\sim$370M \\
\hline
\textsc{T5}~\cite{raffel2020exploring} & Web crawl & ~200B tokens (750GB data) & 11B \\
\hline
\multicolumn{4}{|c|}{(2) \textbf{Multilingual Models}} \\
\hline
\textsc{mBERT}\cite{devlin-etal-2019-bert} & Wiki & 21.9B tokens & 172M \\
  \hline
  \textsc{XLM-R(base)}
\cite{conneau2020unsupervised} & Web crawl & 295B tokens & 270M \\
  \hline
\textsc{XLM-R(large)}
\cite{conneau2020unsupervised} & Web crawl & 295B tokens & 550M \\
  \hline
\textsc{mT5 (large)}~\cite{raffel2020exploring} & Web crawl & 6.3T tokens & 1.2B \\
  \hline
\textsc{mT5 (XXL)}~\cite{raffel2020exploring} & Web crawl & 6.3T tokens & 13B \\
  \hline
\end{tabular}}
    \caption{Training sources, dataset size, and model parameters for popular PLMs. Data sources differ, and are described in the citations listed in each row. 
    }
    \label{tab:plm_size}
\end{table*}

\subsubsection{Autoregressive Language Models} \label{sec:paradigm1-gpt}

An autoregressive language model is trained to predict the next word $x_i$ given all previous words $x_1$, $x_2$, ..., and $x_{i-1}$. The training objective is to maximize the log-likelihood $\sum_i log(P(x_i|x_1, x_2, ..., x_{i-1}); \theta_{T})$, in which $\theta_{T}$ are the model parameters. In a Transformer decoder, these are in multiple layers of multi-head self-attention modules. Typical models include GPT~\cite{radford2018improving}, GPT-2~\cite{radford2019language} and GPT-3~\cite{NEURIPS2020_1457c0d6}\footnote{Open-source re-implementations of GPT are also available, such as GPT-Neo \citep{black2021gptneo} and GPT-J \citep{wang2021gptj}, trained on an 800GB open-source dataset \citep{gao2020pile}, with model sizes similar to GPT-2 (2.7B and 6B parameters respectively).}. 

GPT only utilizes the autoregressive {\it decoder} portion of the Transformer architecture, stacking multiple transformer decoder layers with masked self-attention. This allows the model to attend to all previous tokens in the sequence when predicting the next token. Each newer version of GPT is trained with increasingly large amounts of text (Table~\ref{tab:plm_size}). 


The GPT paper \citep{radford2018improving} proposed fine-tuning GPT for specific tasks, providing examples for natural language inference, QA (including commonsense reasoning), semantic similarity and paraphrase detection, sentiment analysis, and linguistic acceptability (CoLA, \citealp{warstadt2019cola}), 
as well as the GLUE benchmark. In particular, GPT achieves a dramatic improvement on CoLA (scoring 45.4 compared to the previous state of the art of 35.0), showcasing the model's ability to gain a much more sophisticated grasp of language than previous models.
Subsequent versions of GPT (GPT-2 and GPT-3, \citealp{radford2019language, NEURIPS2020_1457c0d6}), however, do not opt for the fine-tuning approach and instead leverage GPT's generative design to tackle tasks in a prompt-based manner or via outright language generation, as described in Sections \ref{sec:paradigm2} and \ref{sec:paradigm3}.

\subsubsection{Masked Language Models} \label{sec:paradigm1-bert}

Whereas autoregressive models are unidirectional, masked language models (MLMs), predict a ``masked'' word conditioned on all other words in the sequence. When training an MLM, words are chosen at random to be masked, using a special token \texttt{[MASK]}, or replaced by a random token.  This forces the model to collect bidirectional information in making predictions. The training objective is to recover the original tokens at the masked positions: $\sum_i m_i log(P(x_i|x_1,...,x_{i-1}, x_{i+1}, ..., x_n); \theta_{T})$, in which $m_i \in \{0,1\}$ indicates whether $x_i$ is masked or not, and $\theta_{T}$ are the parameters in a Transformer encoder. Note that in BERT and similar models, it is a common practice to mask multiple words from a sequence to allow parallel training. Popular examples of MLMs include BERT \citep{devlin-etal-2019-bert}, RoBERTa \citep{liu2019roberta}, and XLM-R \citep{conneau2020unsupervised}.

Specifically, MLMs such as BERT use the {\it encoder} portion of the Transformer architecture. Like autoregressive models, MLMs stack multiple transformer encoder layers to learn increasingly complex and meaningful representations, but it uses masked self-attention to attend to all other tokens in the sequence in both directions when learning a representation for a particular token. The non-autoregressive nature allows the computation to be parallelized, so it is often more efficient at inference time. Dynamic unfolding of all positions in relation to the masked word provides efficiency at training time. 

There is a large family of models derived from BERT, including RoBERTa~\citep{liu2019roberta}, which improves BERT's pre-training, ALBERT~\citep{lan2020albert}, which is smaller and faster to train, and XLNet~\citep{yang2019xlnet} and Transformer-XL~\citep{dai2019transformerxl}, which incorporate an autoregressive pre-training approach to better handle long-distance dependencies. There are also a range of derived models trained on specific domains (Table \ref{tab:domain_plm} in Appendix~\ref{app:plms}). See \citet{qiu2020pre-trained} for a full taxonomy of BERT-derived models.

\subsubsection{Encoder-Decoder Language Models}

The encoder-decoder model is a more flexible ``text in, text out'' model that learns to generate a sequence of token $y_1, ..., y_n$ given an input sequence $x_1, ..., x_m$. Given a pair of sequences, the training objective is to maximize the log-likelihood of $log(P(y_1, ..., y_n|x_1, ..., x_m); \theta_{T})$, in which $\theta_{T}$ are the parameters in a full encoder-decoder Transformer model~\cite{vaswani2017attention}. 

To generate adequate data for self-supervised pre-training, researchers experiment with different forms of sequence corruption. The input is a token sequence modified in some particular way, and the output is the reconstructed, original sequence. Forms of sequence corruption include document rotation, shown in Figure~\ref{fig:plm_arch}, sentence permutation, text infilling, token deletion/masking, and others. Representative models include BART~\cite{lewis-etal-2020-bart} and T5~\cite{raffel2020exploring}. 

Given the sequence-to-sequence (seq2seq) nature, it is straightforward to fine-tune the encoder-decoder language model to perform seq2seq tasks such as Machine Translation, style transfer, and text summarization. The seq2seq formulation is also versatile: many tasks can be reformulated as ``text in, text out''. We describe those approaches in details in Section~\ref{sec:paradigm3}.

\subsection{Pre-Training Corpora} \label{sec:paradigm1-pretraining}



The pre-training corpus is a primary distinguishing factor between language models. Both the size and the quality (source data characteristics) are important considerations. 
Table~\ref{tab:plm_size} presents the sources and the corpus size used for several popular language models. There is a clear trend of increasing the size of the pre-training corpus as well as increasing the diversity of the data. 
For example, ULMFiT \citep{howard2018universal} is trained on a small, highly pre-processed corpus of $\sim$29,000 Wikipedia articles (103 million words), and is representative of models of that year. A few years later, models such as XLM-R \citep{conneau2020unsupervised} and GPT-3 \citep{NEURIPS2020_1457c0d6} leveraged billions of words of crawled web data (diverse in nature).
\citet{raffel2020exploring} observe that the primary gains in performance are typically driven by model size and dataset size (``the bigger, the better''), if the quality of the dataset is held constant. They find that quality can play a larger role if there is a genre match to the task, but a larger dataset provides more advantages, eventually overcoming any gain from quality. For a detailed discussion of model performance scaling by model size, dataset size, and other factors, see \citet{kaplan2020scaling}.
Despite the advantages of the larger dataset, \citet{raffel2020exploring} also demonstrate the importance of cleaning large crawled datasets. They show that a model trained on such an unfiltered dataset performs substantially worse than if filtering heuristics are applied. 
Similarly, GPT-2 \citep{radford2019language} and GPT-3 \citep{NEURIPS2020_1457c0d6} use heuristics to improve the quality of the training data. However, \citet{hendrycks-etal-2020-pretrained} noted that larger models do not necessarily perform better out of domain. \citet{lin2021truthfulqa} also observe that larger language models (trained on these very diverse sources) are more likely to incorrectly answer questions that some humans would answer incorrectly due to false beliefs or misconceptions, thus mimicking the inaccuracies in their training data. 


The domain of intended downstream applications is an important consideration for pre-training source data selection. Table~\ref{tab:domain_plm} (Appendix~\ref{app:plms}) provides a list of domain-specific pre-trained language models that achieved significantly better performance in the intended domain than general-purpose language models. These models are either trained from scratch or trained with domain-specific text using a general-purpose model as the initialization.

\subsection{Fine-Tuning: Applying PLMs to NLP Tasks} \label{sec:paradigm1-finetuning}

\begin{figure*}
  \centering 
  \includegraphics[scale=0.7]{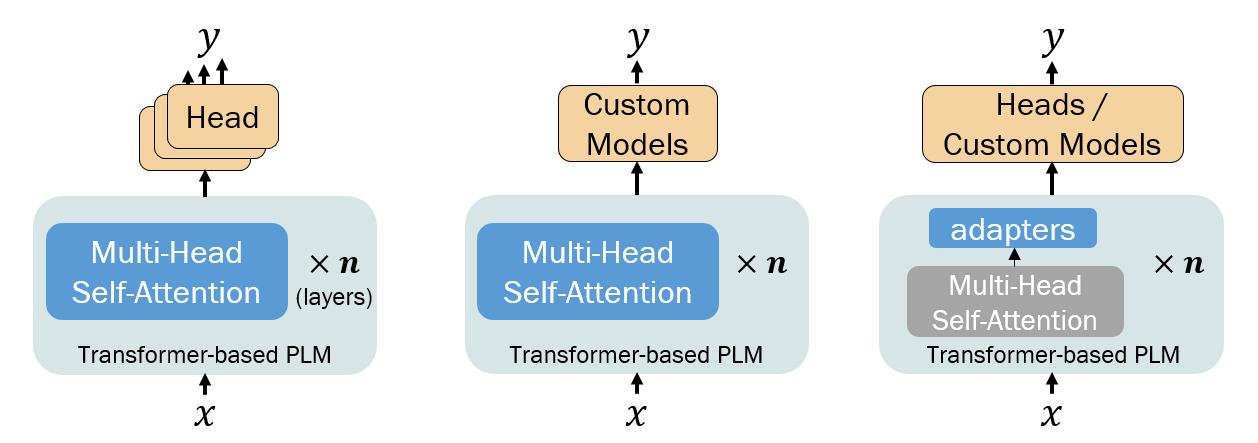}
  \caption{Typical ``pre-train then fine-tune'' strategies. We illustrate strategies that fine-tune the full PLM (left), fine-tune the full PLM in a custom model (center), and fine-tune just a small adapter sub-layer per each Transformer layer (right). We show the Transformer blocks that will be fine-tuned for the specific tasks in blue, and the frozen blocks (keep the pre-trained weights unchanged) in grey. For brevity, we represent the entire Transformer block (stacked in $n$ layers) by its multi-head self-attention and (if applicable) adapter layers. 
  We refer interested readers to \citealp{vaswani2017attention} and \citealp{pfeiffer-etal-2020-adapterhub} for more architecture details. ``Heads'' refers to task-specific prediction functions \cite{wolf-etal-2020-transformers}. }
  \label{fig:pretrain-finetune-archs}
\end{figure*}

Having described the various approaches to creating complex, meaningful representations through pre-training, we turn to the fine-tuning step that allows PLMs to perform accurately on disparate NLP tasks. Figure~\ref{fig:pretrain-finetune-archs} illustrates typical pre-training then fine-tuning strategies. We describe each of them below. A more comprehensive list of prior work using different pre-training then fine-tuning strategies are in Table~\ref{tab:prior_work_by_finetuning_strategy} (Appendix~\ref{app:pre-train-finetune-approaches}). 

\subsubsection{Contextual Embeddings}

The simplest approach to using large pre-trained language models is to ``freeze'' the model and use its output as sophisticated, context-sensitive word embeddings for a subsequent architecture, which is trained from scratch for the specific task. In other words, while this still involves a forward pass through the pre-trained language model over the input text, the language model's weights are {\it not} fine-tuned, rendering this approach closer to a feature extraction family of approaches in classic statistical NLP. There are three types of scenarios for using frozen PLMs. 

In contexts with insufficient labeled data or compute power, ``frozen'' contextual embeddings are employed.  For non-benchmark tasks, the only labeled training datasets are too small
to fine-tune even the top layers of BERT-base, let alone larger models. The computational cost of fine-tuning the entire PLM may be prohibitive for some applications or developers, leading to use of the more efficient frozen PLM solution.  Other data-efficient and time-efficient approaches to fine-tuning are discussed in Section \ref{sec:paradigm1-finetuning-efficient}.

Highly complex or difficult NLP tasks often make use of the frozen PLM technique to help reduce training complexity. Examples are constituency parsing \citep{zhang2020fast}, semantic graph parsing using UCCA\footnote{Universal Conceptual Cognitive Annotation \citep{abend-rappoport-2013-universal}} \citep{jiang-etal-2019-hlt} and AMR\footnote{Abstract Meaning Representation \citep{banarescu-etal-2013-abstract}}  \citep{zhang-etal-2019-amr,naseem-etal-2019-rewarding,zhou-etal-2020-amr}, Aspect-Based Sentiment Analysis \citep{li2019exploiting} and Machine Translation \citep{zhu2020incorporating}. For instance, \citet{zhang2020fast} uses frozen BERT embeddings to seed an innovative approach to Conditional Random Field (CRF) modeling \citep{10.5555/645530.655813} that replaces the inside-outside algorithm with backpropagation, using a two-step process to first bracket then label the parses, and a batched version of the CKY algorithm. For complex tasks like these, there may only be enough data or compute power available to train the secondary model (\citet{zhang-etal-2019-amr} cited limitations in compute power). While the use of frozen PLM parameters is currently in vogue for these tasks, perhaps due to researcher preference for simplicity as well as computational requirements, we may see a shift to full-model fine-tuning for tasks with sufficient training data.


Unsupervised tasks such as word sense disambiguation \citep{hadiwinoto-etal-2019-improved} and word sense induction \citep{amrami2019better} are not associated with a supervised dataset for fine-tuning. Instead, frozen BERT embeddings are fed through a variety of strategies such as nearest-neighbour matching, affine transformations, gated linear units (GLU, \citealp{dauphin2017language}) or clustering algorithms to perform these tasks.  

\subsubsection{Fine-tuning the PLM}

This approach fine-tunes some or all the layers of the PLM and then adds one or two simple output layers (known as prediction heads, \citealp{wolf-etal-2020-transformers}). Typically, these are feed-forward layers for classification. The output layers and the PLM are trained together in an end-to-end setup, but the bulk of the computation is applied to fine-tuning the language model to produce the desired representation of the input. The task of the output layers is merely to condense the information provided by the embeddings of each token into the number of desired classes. The word embeddings may come from the top layer, or from a concatenation or a weighted average of the top $n$ (often $n=4$) layers \citep{peters2018deep}. Figure~\ref{fig:pretrain-finetune-archs} (left) shows an illustration of this approach.

This approach is most suitable for sequence classification tasks (e.g. sentiment analysis, NLI, semantic similarity), sequence tagging tasks such as NER, and span extraction tasks (e.g. QA) in which the newly trained layers learn the start and end span of an answer. 


For sequence classification tasks, \citet{devlin-etal-2019-bert} suggests fine-tuning BERT's representation of the special \texttt{[CLS]} token, and following with a single feed-forward layer that classifies it as one of the task labels. For token-level or span-level classification tasks, the representations of each token, or alternatively just the representation of the first sub-token of each token or span (as in \citealp{devlin-etal-2019-bert}), may be passed to the classifier. This fine-tuning approach is use to apply BERT to all 11 tasks in GLUE, as well as QA (SQuAD), NER (CoNLL 2003), and common-sense inference (SWAG). For many additional examples of this highly popular approach, see Table \ref{tab:prior_work_by_finetuning_strategy} (Appendix ~\ref{app:pre-train-finetune-approaches}).

In this setting, care is needed to choose an appropriate learning rate that works for both the weights of the feed-forward layer(s) and for the PLM. Since the PLM is already largely trained, a low learning rate should be used (between 1e-3 \citep{raffel2020exploring} and 1e-5 \citep{liu2019roberta}), with a lower learning rate for smaller datasets. However, the randomly initialized feed-forward layer weights still require significant training. As such, it is a common practice to freeze the language model layers temporarily while initially training the feed-forward layers, then unfreeze the language model gradually for additional fine-tuning \citep{howard2018universal,yang2019xlnet}. The degree to which this should be done depends on the size of feed-forward layers, and whether a token such as BERT's \texttt{[CLS]} is being used. If the majority of the labour is being done by \texttt{[CLS]}, as in all the examples in \citet{devlin-etal-2019-bert}, there are fewer benefits to training the feed-forward layer alone. Again, this is a function of the availability of supervised training data.

The next choice is how many layers of the PLM to fine-tune. While the examples in the BERT paper fine-tune the entire model, this is not feasible for NLP tasks with small datasets or in situations where compute power is a limitation. Often, tuning just the top few layers of the language model is sufficient; for example, \citet{ross-etal-2020-exploring} only fine-tune the top layer of BERT on their small supervised dataset of 2000 sentences. A range of papers in the growing field of ``BERTology'' (\citealp{tenney-etal-2019-bert}, \citealp{clark-etal-2019-bert}, \citealp{rogers-etal-2020-primer}) show that the lower layers of BERT contain word-specific and syntactic information such as part of speech, while the upper layers contain more semantic and increasingly complex information such as semantic roles and coreference information.

\subsubsection{Fine-tuning the PLM in Customized Models}

Some tasks require significant additional architecture on top of a language model, as illustrated in Figure~\ref{fig:pretrain-finetune-archs} (center). With sufficient training data and computational power, researchers may choose to train both a substantial task-specific architecture and also fine-tune the language model. This is the preferred choice for structure prediction tasks, in particular parsing tasks and occasionally sequence tagging tasks. Examples of sequence tagging models using this approach include BERT-CRF for NER ~\cite{souza2020portuguese,taher2019beheshti}, though notably \citet{devlin-etal-2019-bert} show that the Conditional Random Field (CRF)  layer is not necessarily needed for NER with BERT. Examples of parsing models using this approach include UDapter for dependency parsing \citep{ustun-etal-2020-udapter}.

Any sequence-to-sequence task that uses a pre-trained language model as its encoder may employ this approach. An interesting example is \citet{zhu2020incorporating}'s formulation of machine translation. However, \citeauthor{zhu2020incorporating} did not find any significant improvement over using BERT-based frozen word embeddings. 

A related and highly successful approach is to fine-tune the entire language model with a small number of feed-forward layers,  then layer on an algorithmic approach that provides a substantial amount of task-specific heavy lifting. For example, it might transform the task from a classification problem (as understood by the language model) into the desired target formulation, often a structured form such as a tree or a set of clusters. For coreference resolution, \citet{joshi2019bert,joshi-etal-2020-spanbert} adds a substantial algorithm, in their case e2e-coref \citep{lee-etal-2018-higher} which transforms ratings of pairs of spans into valid mention clusters. Specifically, for each candidate mention span, the algorithm computes a distribution over possible antecedent spans from the mention score (whether it is likely to be a mention) and the compatibility score of the two spans, which itself involves a feed-forward network to compute.
Two more structural parsing examples in this vein are temporal dependency parsing \citep{ross-etal-2020-exploring} and modal dependency parsing \citep{yao2021factuality}. These studies approach tree building algorithmically by first performing a classification problem to identify suitable dependency pairs, then ranking them to construct a valid tree. 

\subsubsection{Efficient Fine-tuning Approaches} \label{sec:paradigm1-finetuning-efficient}

A wide range of approaches, in addition to limiting fine-tuning to the top layers, seek to fine-tune only a small number of model weights. These can be classified into two types: (a) fine-tuning a separate, small network that is tightly coupled with the PLM (but does not change it), and (b) selecting only a small number of the PLM's weights to fine-tune or keep.

The most prominent approach of the first type are adapter modules \citep{houlsby2019parameter,bapna2019simple,pfeiffer-etal-2020-mad, pfeiffer-etal-2020-adapterhub}, as illustrated in Figure~\ref{fig:pretrain-finetune-archs} (right). Adapters add a small set of newly
initialized weights at every layer of the transformer. \citet{houlsby2019parameter} show that a two-layer feed-forward network with a bottleneck works well. The  placement and configuration of the adapters within the Transformer blocks varies in the literature \citep{houlsby2019parameter,bapna2019simple,stickland2019bert,pfeiffer-etal-2020-mad}. During fine-tuning, all weights in the PLM remain frozen except for the few weights in the adapters. One set of adapters is fine-tuned per task of interest. This approach is more efficient in training (typically $<5\%$ of all PLM weights), and allows efficient weight-sharing, both in terms of using the same frozen PLM for each task, and in allowing the weights of adapter modules to be distributed and also re-used. Notably, the weights of adapters independently trained for different tasks can be successfully combined to solve a new task \citep{pfeiffer-etal-2020-mad}. Finally, catastrophic forgetting of old capabilities when fine-tuning on a new task or language is prevented. 
AdapterHub \citep{pfeiffer-etal-2020-adapterhub} and Trankit \citep{nguyen2021trankit} are examples of frameworks promoting an adapter ecosystem; an example of using adapters for Universal Dependency Parsing is \citet{ustun-etal-2020-udapter}. 

A similar method is side-tuning \citep{zhang2020sidetuning}, which adapts a pre-trained network by training a lightweight ``side'' network that is fused with the (unchanged) pre-trained network using a simple additive process. 
Also closely related is diff-pruning \citep{guo2021parameterefficient}, which adds a sparse, task-specific difference vector to the original (frozen) parameters. These difference vectors are regularized to be sparse, which further decreases the number of weights that need to be stored (around 0.5\% of the original model's parameters). 

Moving to the second type of approach, BitFit~\cite{zaken2021bitfit} proposes to limit  fine-tuning to the bias terms (or a subset of the bias terms, around 0.1\% of the total parameters) of pre-trained BERT models, plus a task-specific classification layer. This is shown to be competitive with, and for some tasks better than, fine-tuning all of BERT. BitFit builds on the intuition that fine-tuning exposes existing capabilities, rather than teaching the model new ones.

Similarly, \citet{pmlr-v108-radiya-dixit20a} show that it suffices to fine-tune only the ``most sensitive'' layers, i.e. those which are most distant in parameter space from the rest of the model. In parallel, they sparsify the model substantially by 
setting 1-4\% of pre-trained parameters to zero. This retains performance, as also demonstrated by work like DistilBERT \citep{sanh2020distilbert} and other pruning studies (\citealp{prasanna-etal-2020-bert} inter alia) which show that many parameters in a large PLM are redundant.

In fact, \citet{zhao-etal-2020-masking} propose masking, i.e. setting weights to zero, as a sole alternative to fine-tuning the model weights. This approach freezes all the weights of the PLM,  selects the weights that are relevant for a given task, and  masks (discards) the rest. They train one mask per downstream task, with every layer masked except the embedding layer. While in principle this trains as many parameters as the original model, the mask is both binary and sparse and thus much simpler to learn and store. The initial sparsity of the mask is an important hyperparameter in this approach, as is deciding which layers to mask, since the different layers encode various degrees of syntactic and semantic knowledge \citep{tenney-etal-2019-bert}. \citeauthor{zhao-etal-2020-masking} show that masking ``top-down'' (mostly the top layers, which are more task-specific and encode more semantic and long-distance information) is more effective than masking ``bottom-up'' (which would mask mostly the layers dealing with elementary word meaning and syntax). In particular, performance on CoLA increases as more layers are masked top-down. The authors further show that masking yields entirely comparable performance to fine-tuning on a range of tasks from POS tagging to reading comprehension.

\section{Paradigm 2: Prompt-based Learning} \label{sec:paradigm2}

\begin{figure*}
    \centering
    \resizebox{\textwidth}{!}{
        \includegraphics{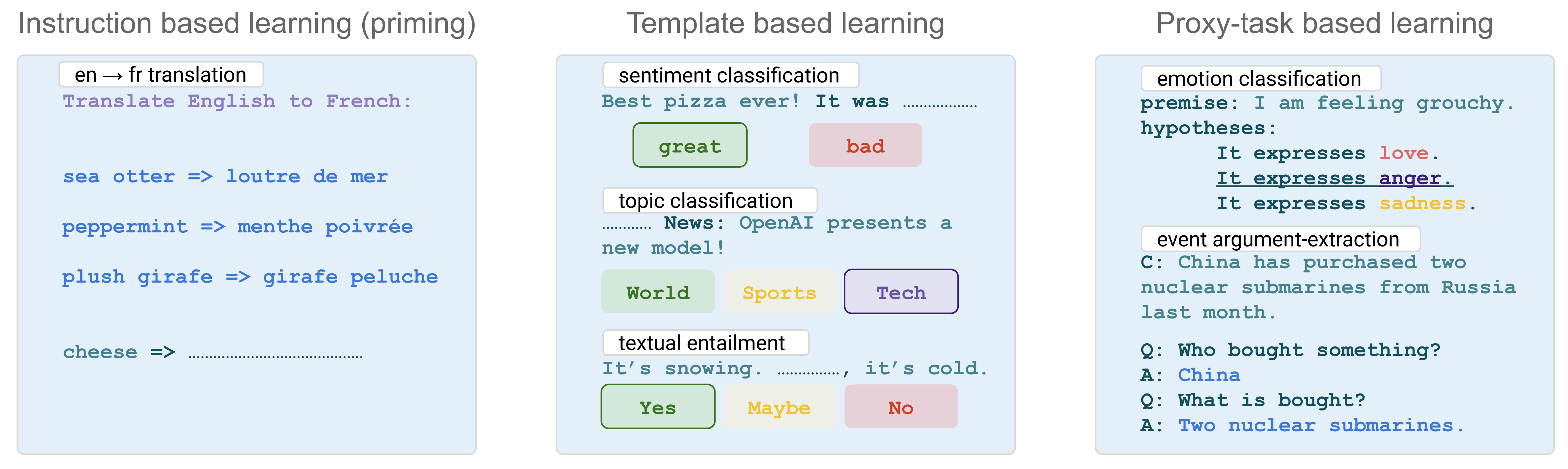}
    }
    \caption{The three main prompt-based approaches. On the instruction based learning (left box) the instructions are marked in purple, the in-context examples in blue and the prompt in cyan. On the prompt based learning (middle box), the text to classify is marked on light cyan and the prompt on dark cyan; the label verbalizations are shown in small boxes. On the proxy task based learning (right box), prompts are marked with dark cyan, the context is on light cyan and the answers generated by the model are in blue.}
    \label{fig:prompt-methods}
\end{figure*}


We use prompting to refer to the practice of adding natural language text, often short phrases, to the input or output to encourage pre-trained models to perform specific tasks \cite{yuan2021bartscore}.
There are several advantages to using prompts. Prompting, especially in-context learning (e.g. ~\citealp{NEURIPS2020_1457c0d6}), may not require updates to the PLM's parameters, reducing computational requirements as compared to fine-tuning approaches, or in addition to those described in ~\ref{sec:paradigm1-finetuning-efficient}.  Prompts also encourage a better alignment of the new task formulation with the pre-training objective, leading to better use of knowledge captured in pre-training.  The closer match also enables a few-shot approach \citep{liu2021pretrain}, especially for tasks with small training datasets;  a good prompt can be worth hundreds of labeled data points \citep{lescao2021many}. Finally, prompts allow probing of the PLMs, often in an unsupervised way, in order to assess the knowledge acquired by the PLM for specific tasks of interest \citep[e.g.][]{petroni2019language}.

We discuss 3 types of prompt-based learning approaches below: Learning from instructions and demonstrations, template-based learning, and learning from proxy tasks. Figure 3 shows illustrations for each of the three approaches.


\subsection{Learning from Instructions and Demonstrations} \label{subsec:instruction_learning}

First attempts made use of instructions such as ``translate X to Y:'' \citep{raffel2020exploring} to simultaneously teach the model varied tasks in a text-to-text manner. However, this approach required a large amount of labeled data.

With the emergence of large generative PLMs \citep{radford2019language}, the first signs that language models are multi-task learners emerged. For instance, GPT-2 understands that if the instruction ``TL;DR'' (``too long; didn't read'') is given, then it should generate a summary of the context following the instruction. More recently, and with even larger generative PLMs (GPT-3), \citet{NEURIPS2020_1457c0d6} showed that those models are indeed very good at few-shot learning. \citeauthor{NEURIPS2020_1457c0d6} showed that GPT-3 can perform few-shot tasks via priming (in-context learning): given instructions and a few input/output pairs, GPT-3 is able to produce the desired outputs for new inputs. No gradient updates are performed (see Figure \ref{fig:prompt-methods}, left box). Caveats include the requirement of a very large LM to work well, and an inability to scale to more than a few examples, because the context window of most LMs is limited to a few hundred tokens. 
We refer readers to the GPT-3 paper~\cite{NEURIPS2020_1457c0d6} for many additional examples of learning from instructions and/or demonstrations.


\begin{table}[]
    \centering
    \scalebox{0.75}{
\begin{tabular}{|p{0.55\linewidth}|p{0.65\linewidth}|}
\hline
\textbf{Input} & \textbf{Output} \\
\hline
\multicolumn{2}{|c|}{{\bf (1) Text pair generation~\cite{schick2021generating} }} \\
\hline
Task: Write two sentences that mean the same thing. Sentence 1: ``A man is playing a flute.'' Sentence 2: \rule{0.3cm}{0.15mm}	& ``He's playing a flute.'' \\
\hline
\multicolumn{2}{|c|}{{\bf (2) Mathematical reasoning~\cite{reynolds2021prompt} }} \\
\hline
$f(x) = x*x$. What is $f(f(3))$? Let's solve this problem by splitting it into steps. \rule{0.3cm}{0.15mm} & $f(f(3)) = f(3*3) = 3*3*3 = 27$. We can see that $f(3) = 3*3 = 9$, so $f(f(3)) = 27$. \\
\hline
\end{tabular}}
    \caption{Example prompt designs for learning from instructions.}
    \label{tab:prompting_with_instructions}
\end{table}

\citet{schick2021generating} and \citet{reynolds2021prompt} introduce new tasks based on descriptions.  For example, the text pair generation task \citet{schick2021generating} consists in generating a continuation sentence (Sentence 2) given an input sentence (Sentence 1) and a description of the relations between the sentences (Table~\ref{tab:prompting_with_instructions}(1)).\footnote{A variation of this task consists in generating both Sentence 1 and Sentence 2 given the description \cite{schick2021generating}.} To address this task, \citet{schick2021generating} use a generative PLM (GPT2-XL) that generates Sentence 2, replacing the  \rule{0.3cm}{0.15mm} token.  Impressively, even mathematical reasoning can be handled (Table~\ref{tab:prompting_with_instructions}(2)): \citet{reynolds2021prompt} show that by inserting a natural language prompt (``Let's solve \ldots steps.'') after the math problem statement, GPT-3 can generate a procedure that solves the math problem. 

Recently, \citet{wei2021finetuned} showed that teaching a very large PLM to follow instructions with supervised data improves the zero and few-shot abilities of these PLMs. They carried out a large scale multi-task experiment over more than 60 datasets grouped into 12 task different tasks, and showed that a PLM trained via natural language instructions on other tasks outperforms a standard language model on the test task. \citet{mishra2021crosstask} fine-tuned BART \citep{lewis-etal-2020-bart} to perform a similar task using instructions and few-shot examples for a variety of crowd-sourced NLP tasks. The crowdsourcing process of each task consists of several steps that are natural and intuitive for human annotators. The instructions to the PLM match the step-by-step crowdsourcing instructions, decomposed into self-contained, separate tasks, leading to improved performance on unseen tasks, in contrast to an earlier work \citep{efrat2020turking} that reported negative performance when using the crowd-sourcing instructions as-is.

Scaling limitations may affect the broad applicability of this approach: \citet{wei2021finetuned} show that instruction tuning achieves significant improvements on held-out tasks in the zero-shot setting when using very large PLMs (e.g.~with 68B or 137B parameters), but hurts performance when applied to PLMs with 10B parameters or less. In a similar setting, \citet{sanh2021multitask} showed that it is possible for a model with 11B parameters to benefit from instruction tuning, and  identified three key differences compared to \citet{wei2021finetuned}. (1) They use a encoder-decoder model trained first with the MLM objective, then as a standard LM, and finally fine-tuned on a multitask objective, rather than a decoder-only autoregressive LM. (2) They argue that their prompts are qualitatively more diverse in terms of length and creativity. (3) They hold out multiple tasks at once, rather than only one at a time. 


We note that the descriptions in instruction learning can be very detailed.
For example, the crowd-sourcing instructions in  \citet{mishra2021crosstask} contain the task definition, things to avoid, emphasis and caution (i.e. required properties for the output), and positive and negative examples.

\subsection{Template-based Learning} \label{subsec:template-based_learning}

A more widely used approach, template-based learning, reformulates NLP tasks into tasks that are closer to language models' pre-training tasks via template-based prompts. This better leverages the knowledge captured in the pre-training tasks, leading to a significant reduction in the number of task-specific training examples required to achieve a similar performance to previous approaches\citep{lescao2021many}, or even eliminating the need for training data. To achieve this goal, template-based learning reformulates various NLP tasks into language modeling tasks via carefully designed templates with open slots. In this way, solving the tasks is reduced to filling the slots with words or phrases using PLMs, and then projecting these outputs into the task-specific labels. 

Template-based learning differs from instruction learning (Section \ref{subsec:instruction_learning}) in that templates are less detailed and do not explicitly describe the task.

\subsubsection{Template Design} \label{sec:template design} 

\begin{table}[]
    \centering
    \scalebox{0.75}{
\begin{tabular}{|p{0.55\linewidth}|p{0.65\linewidth}|}
\hline
\textbf{Input} & \textbf{Output} \\
\hline
\multicolumn{2}{|c|}{{\bf (1) Topic/sentiment classification~\cite{schick-schutze-2021-exploiting}}} \\
\hline
{\it Best pizza ever!. It was \rule{0.3cm}{0.15mm}.} & {\it great $\rightarrow$ Positive} \\
\hline
\multicolumn{2}{|c|}{{\bf (2) Textual entailment~\cite{schick-schutze-2021-exploiting}}} \\
\hline
{\it Mia likes pie? \rule{0.3cm}{0.15mm}, Mia hates pie.} & {\it No $\rightarrow$ Contradiction} \\
\hline
\multicolumn{2}{|c|}{{\bf (3) Event argument extraction~\cite{chen-etal-2020-reading}}} \\
\hline
{\it Americans sought to bring calm to Mosul, where U.S. troops killed 17 people in clashes earlier in the week.} \underline{someone} killed \underline{someone} with \underline{something} in \underline{some place} at \underline{some time}. & \underline{U.S. troops} killed \underline{17 people} with \underline{something} in \underline{Mosul} at \underline{earlier in the week}.  \\
\hline
\multicolumn{2}{|c|}{{\bf (4) Probing for relations/facts~\cite{petroni2019language}}} \\
\hline
{\it Dante was born in \rule{0.3cm}{0.15mm}.} & {\it Florence}\\
\hline
\multicolumn{2}{|c|}{{\bf (5) Probing for commonsense~\cite{trinh2019simple}}} \\
\hline
{\it The trophy doesn’t fit in the suitcase because \underline{it} is too big} & ${\it it \rightarrow trophy}$:0.9 ${\it it \rightarrow suitcase}$:0.2 \\
\hline
\multicolumn{2}{|c|}{{\bf (6) Probing for reasoning~\cite{talmor2020olmpics}}} \\
\hline
{\it The size of an airplane is \rule{0.3cm}{0.15mm}  than the size of a house . A. larger B. smaller} & {\it larger} \\
\hline
\end{tabular}}
    \caption{Example prompt designs for template-based methods. {\it great $\rightarrow$ Positive} means that the answer {\it great} will be converted to label {\it Positive}. For  \citet{chen-etal-2020-reading}, each of the underlined words (e.g. {\it \underline{someone}}) will be replaced with the underlined phrase on the output side by a PLM. ``{\it it $\rightarrow$ trophy}: 0.9'' means by replacing the underlined pronoun {\it it} with {\it trophy}, the modified sentence has a likelihood score of 0.9 according to the PLM. }
    \label{tab:prompt_with_template}
\end{table}

Using a {\it cloze-style prompt} design, inputs to a PLM are converted into a format such that solving the NLP task only requires the PLM to predict missing word(s). Table~\ref{tab:prompt_with_template} (1) shows the most straightforward example of this approach, as applied in the sentiment detection domain. 

For classification tasks, each predicted word or phrase is converted into a class label of interest. For example, we can design a cloze-style prompt for a textual entailment task in which the goal is to predict the {\it entail/contradict} relation between a pair of input sentences $ \langle X_1, X_2 \rangle$. Pattern-Exploiting Training (PET) \citep{schick-schutze-2021-exploiting} (Table~\ref{tab:prompt_with_template} (2)) converts a pair of inputs $\langle X_1, X_2 \rangle$ into  ``$X_1 {\rm ? \rule{0.3cm}{0.15mm}}, X_2$'' and asks a masked language model to predict the missing word. The prediction (here {\it yes} or {\it no}) is  directly mapped to one of the textual entailment class labels. This template design allows PET to reformulate the text entailment problem into the same masked language modeling problem that was used
to pre-train the PLM.  Therefore, it is popular among classification tasks that may be reformulated as predicting a masked word or short phrase (e.g. topic classification, textual entailment, and knowledge probing).
\citet{chen-etal-2020-reading} reformulates the event argument extraction challenge as a cloze-style problem (Table~\ref{tab:prompt_with_template} (3)), predicting fillers for the underlined positions, and then apply greedy decoding to fill in the \rule{0.3cm}{0.15mm} position incrementally. \citet{petroni2019language} similarly use the cloze-style task for relation/fact probing (Table~\ref{tab:prompt_with_template} (4))

A {\it multiple-choice style prompt} proves useful for probing for commonsense knowledge. This kind of template provides a selection of hypotheses for the PLM, which selects its preferred answer. For example, in Table~\ref{tab:prompt_with_template} (5), \citet{trinh2019simple}'s model selects {\it trophy} instead of {\it suitcase} to replace \underline{it} in the original sentence.  Table~\ref{tab:prompt_with_template} (6) shows work by \citet{talmor2020olmpics}, expressing similar reasoning through a hypothesis-driven approach.

{\it Prefix prompts} \citep{li2021prefixtuning,hambardzumyan-etal-2021-warp,lester2021power} are another common type of template. Prefixes are task-specific vectors prepended to the input. They do not correspond to actual words but consist of free parameters. Prefix prompts are usually the best choice for tasks that require generating text or predicting a next word or phrase, because the prefix-based prompt design is consistent with the left-to-right nature of the autoregresive model \citep{liu2021pretrain}.

Prompts can be further augmented via adding demonstrations ({\it demonstration learning}) \cite{gao-etal-2021-making}. In that case, a few labeled examples are appended to the template to make it more informative, allowing the PLMs to better predicting the answer. 

\subsubsection{Template Construction}

Templates can be either manually crafted or automatically generated. We here survey the different methods for generating them as well as ways to combine and manipulate the template-based prompts (multi-prompt learning).

\paragraph{Manually-crafted Templates.}

Most early work in prompt-based learning uses some form of manually crafted templates. For example, manual cloze templates are used in \citet{petroni2019language} to probe the knowledge of the model, as well as in \citet{schick-schutze-2020-bertram}, \citet{schick-etal-2020-automatically} and \citet{schick-schutze-2021-exploiting} for text classification in a few-shot setting. Manually designed prefix prompts are leveraged in \citet{NEURIPS2020_1457c0d6} for QA, translation, and probing tasks for commonsense reasoning. 
The quality of the prompts impacts performance. Indeed, \citet{zhao2021calibrate} showed that different prompts can cause accuracy to vary from near chance to near state-of-the-art.

\paragraph{Automatically-generated Discrete Templates.} Discrete templates, which usually correspond to natural language phrases, are described in a discrete space. To search for such templates given a set of inputs and outputs, \citet{jiang2021know} proposed a mining-based approach called MINE that aims to find either the middle words or dependency paths between the inputs and outputs. A second approach \citep{jiang2021know,yuan2021bartscore} consists of paraphrasing an existing template prompt using back and forth machine translation, and then selecting the best prompt among the new paraphrases with guidance from a thesaurus.  Prompt paraphrasing is also used by \citet{haviv-etal-2021-bertese} who used a neural prompt rewriter that optimizes the accuracy of systems using the prompt. In that case, a different paraphrase is generated for each input.  A third approach uses gradient-based search to find short sequences that can serve as prompt \cite{wallace-etal-2019-universal,shin-etal-2020-autoprompt}.
\citet{gao-etal-2021-making} and \citet{bendavid2021pada} further generate prompts using standard generation models such as T5 \cite{raffel2020exploring}. In the latter, the authors proposed a domain
adaptation algorithm that trains T5 to generate unique domain relevant features that can be concatenated with the input to form a template for downstream tasks.

\paragraph{Automatically-generated Continuous Templates.} Continuous prompts, which perform prompting directly in the embedding space of the model, allow us to abstract away from natural language prompts (i.e. the prompts do not correspond to actual words) and from the parameters of the LM \citep{liu2021pretrain}. These continuous prompts often require tuning on task-specific data. \citet{li2021prefixtuning} propose prefix tuning, which prepends a sequence of continuous, task-specific vectors to the input while keeping the LM parameters frozen. This allows them to fine-tune just 0.1\% of the total model parameters. A similar method is used by \citet{lester2021power}, who differ from \citet{li2021prefixtuning} by adding special tokens to form a template and tuning the embeddings of these tokens directly, without introducing additional
tunable parameters within each network layer. Continuous prefix tuning is also used by \citet{tsimpoukelli2021multimodal} in the context of multimodal learning (language and vision) but in that case the prefix is sample dependent. Tuning can be initialized with discrete prompts as in \citet{zhong-etal-2021-factual}, \citet{qin-eisner-2021-learning} and \citet{hambardzumyan-etal-2021-warp}. It can also be done by inserting some tunable embeddings into a hard prompt template as in \citet{liu2021gpt} and  \citet{han2021ptr}, who propose prompt tuning with rules (PTR). This uses manually crafted sub-templates to compose a complete template using logic rules (see Section \ref{sec:template-applications} for its application to relation extraction).

It is worth noting that \citet{logan2021cutting} showed that fine-tuning PLMs in the few-shot setting can avoid prompt engineering, and that one can use prompts that contain neither task-specific templates nor training examples, and even {\it null prompts} that are simple concatenations of the inputs and
the [MASK] token and still achieve competitive accuracy on NLU tasks.

\paragraph{Multi-Prompt Learning} A number of approaches use prompt ensembling, augmentation, and decomposition/composition for a more flexible task design. We describe them below.

First, multiple prompts can be used for an input (dubbed {\it prompt ensembling}) at inference time. The prompts can be combined using a uniform average \cite{jiang2021know,schick-schutze-2021-exploiting,yuan2021bartscore} or a weighted average \cite{jiang2021know,qin-eisner-2021-learning,schick-schutze-2021-exploiting,schick2021just}. Another way to combine the prompts is majority voting to combine the results of the different prompts as in \citet{lester2021power} and \citet{hambardzumyan-etal-2021-warp}. Knowledge distillation  \cite{allenzhu2021understanding}, where the idea is that the knowledge present in an ensemble of models can be distilled into a single model, has been borrowed to the context of prompt combination by \citet{schick-schutze-2021-exploiting,schick2021just,schick2020fewshot} and \citet{gao-etal-2021-making} where for each  template-answer pair a separate model is trained, before ensembling them to annotate an unlabeled dataset. Then, the authors train a new model to distill the knowledge from the annotated dataset. In the case of generation tasks, \citet{schick2020fewshot} trained a separate model for each prompt. Then the model outputs were scored by averaging their generation probability across all models.

Second, prompts can be decomposed or composed to more effectively solve an NLP task. Decomposition involves finding sub-problems for which prompts can be generated separately. For example, \citet{cui2021template} proposed an approach for named entity recognition, where the different prompts for each candidate span were created and predicted separately.

Third, augmentation methods such as {\it demonstration learning} \cite{gao-etal-2021-making} create more descriptive prompts, as in a multiple-choice problem. \citet{lu2021fantastically} showed that both the choice of examples in the prompts and the order of the prompts can considerably affect the results. To select the examples from which the PLM must choose the correct response ({\it example sampling}),  \citet{gao-etal-2021-making} and \citet{liu2021makes} used sentence embeddings to find examples semantically close to the input. \citet{mishra2021crosstask} used both positive and negative examples, teaching the PLM types of items to avoid in performing new tasks with only instructions. As for the order of the selected examples ({\it sample ordering}), \citet{kumar-talukdar-2021-reordering} searched for the best permutation of prompts and also learned a segmentation token to separate between the prompts.
They showed the usefulness of this method for few-shot learning on the task of sentiment classification.

\subsubsection{Answer Generation} \label{sec:answer-generation}

There are two main types of answers to prompts: those that map to a classification label (e.g. \citealp{yin-etal-2019-benchmarking,cui2021template}), and those intended as the final answer (e.g. \citealp{petroni2019language,jiang-etal-2020-x,radford2019language}). For classification tasks, typically addressed with cloze-style prompts, the developers identify a subset of words and phrases from which the PLM may choose, and that choice is easily mapped to the class of interest. For instance, in a sentiment detection task, the PLM may answer a prompt with ``good,'' ``great,'' or ``excellent,'' all of which are mapped to a ``positive'' sentiment label. The second type of answer, free text, prevails for text generation tasks. Examples of both types are shown in Table~\ref{tab:prompt_with_template}.



In either case, the definition of the answer space may be optimized to produce ideal prompt responses. \citet{jiang2021know} used {\it paraphrasing} to extend the search space with back translation (translating to another language,  then back to the original). Another approach, explored by \citet{schick-schutze-2021-exploiting}, \citet{schick-etal-2020-automatically}, \citet{shin-etal-2020-autoprompt} and \citet{gao-etal-2021-making}, is {\it prune-then-search}, a two-step method where the answer space is pruned, for example by only selecting
a subset of words according to their zero-shot accuracy on the training data \cite{gao-etal-2021-making} and then an answer is searched in the pruned space. An approach called {\it label decomposition} optimizes the search space by modeling the label names for comparison to the answer tokens; for instance, in \citet{chen2021knowprompt} the decomposed relation labels (their individual tokens) represent the answer space. Lastly, \citet{hambardzumyan-etal-2021-warp} add a virtual token for each class label and optimize its embedding together with the token embeddings of the prompts, using gradient descent. This {\it gradient descent optimization} approach allows direct optimization of the answers instead of using a discrete search.


\subsubsection{Task-specific Tuning} \label{sec:task-tuning}


While prompts can be directly used in a zero-shot, unsupervised setting, prompts have also been used in fully supervised or few-shot settings where either all or part of the specific-task training data is available. Two main approaches currently prevail for tuning a PLM with prompts.

The first approach uses a fixed template-style prompt to perform tuning of the PLM. Here, a fixed template is usually applied to every training and test example as in the PET-TC \citep{schick-schutze-2021-exploiting}, PET-Gen \citep{schick2020fewshot}  and  LM-BFF \citep{gao-etal-2021-making} models. \citet{lescao2021many} quantified the benefit of using prompts in classification tasks by fine-tuning in equal conditions across many tasks and data sizes. They showed that prompting consistently improves the results across tasks over just fine-tuning, that it is most robust to the choice of pattern, and that it can be learned without an informative verbalizer (a function that maps
each label to a single vocabulary token). \citet{logan2021cutting} showed that only tuning 0.1\% of the parameters in the prompt-based few-shot setting can achieve comparable or better accuracy than standard fine-tuning. For this purpose, they explored different ways to perform memory-efficient fine-tuning, including (i) Adapters \citep{houlsby2019parameter}, which are neural network layers inserted between the feed-forward portion of the Transformer architecture (see Section \ref{sec:paradigm1-finetuning-efficient}); (ii) BitFit \citep{zaken2021bitfit}, where
only the bias terms inside the Transformer are updated; (iii) PLM head tuning, where the embeddings in the MLM output layer that are associated with the tokens of the verbalizer  are updated; and (iv) Calibration \citep{zhao2021calibrate}, where an affine transformation on top of the logits associated with the verbalizer tokens is learned. They found that the best results are achieved using BitFit.

The second approach is joint tuning of the prompt and the PLM. Here, prompt-relevant parameters are fine-tuned together with the all or some of the
parameters of the PLM, as in  PADA \citep{bendavid2021pada}, where the prompts are properties of source domains, generated based on their relatedness to the input example (from a new domain), and P-Tuning \citep{liu2021gpt}, which makes use of trainable continuous prompt embeddings when applying GPT models on NLU tasks. Finetuning both the model and the prompt-relevant parameters makes this approach very expressive. On the other hand, it requires the storage of all the parameters, with makes it less applicable to small datasets \cite{liu2021pretrain}.

It is worth noting that task-specific training can also be used earlier during the construction and  validation of the prompts.
Indeed, as pointed out by \citet{perez2021true}, previous PLM-based few-shot learning approaches used many held-out examples to tune various aspects of learning, such as hyperparameters, training objectives, and natural language templates (``prompts''). 
 \citet{perez2021true} propose instead to evaluate the few-shot ability of PLMs in a {\it true few-shot learning} setting, where such held-out examples are unavailable. 

\subsubsection{Applications of Template-based Methods} \label{sec:template-applications}

Template-based prompting methods are currently applied to a growing list of NLP tasks. We provide a survey of how recent studies have addressed a varied set of NLP applications.

\paragraph{Text Classification.} In \citet{puri2019zeroshot}, natural language descriptions of classification tasks were given as input. Then, the model was trained to generate the correct answer in natural language via a language modeling objective, aiming to generalize to new classification tasks without task-specific tuning. 

\paragraph{Information Extraction (IE).} \citet{cui2021template} considered the NER task as a language model ranking problem in a sequence-to-sequence framework where the source sequence corresponds to the original sentence and the target sequence corresponds to the template prompt, 
filled by candidate spans. 
For the relation extraction task, \citet{han2021ptr} proposed a model called Prompt Tuning with Rules (PTR), which applies logic rules to construct prompts
with several sub-prompts. \citet{chen2021knowprompt}, instead of using rules, constructed the prompts by leveraging learnable virtual template words and virtual answer words. Their representation is synergistically optimized with knowledge constraints. For the event extraction task in a cross-lingual setting, \citet{fincke2021language} proposed using the event type and an integer representing the argument type as prefixes.

\paragraph{Knowledge Probing.} Factual probing has been explored in particular by \citet{petroni2019language} and \citet{jiang-etal-2020-x} to quantify the amount of factual knowledge already present in the PLMs, providing the LAMA and X-FACTR datasets, respectively. Other works that investigated model knowledge with discrete template search  include \citet{petroni2020context}, \citet{jiang2021know}, \citet{haviv-etal-2021-bertese}, \citet{shin-etal-2020-autoprompt} and \citet{perez2021true}. Continuous template learning was used in \citet{qin-eisner-2021-learning}, \citet{liu2021gpt} and \citet{zhong-etal-2021-factual}. Prompt ensemble learning was applied to knowledge probing by \citet{jiang2021know} and \citet{qin-eisner-2021-learning}. 

In addition to factual knowledge, additional types of knowledge that have been probed using the cloze test include commonsense ~\citep{trinh2019simple}, relational knowledge ~\citep{petroni2019language}, reasoning ~\citep{talmor2020olmpics} and understanding rare words~\citep{schick2019rare}.
For commonsense reasoning, Winograd Schemas \citep{levesque2012winograd} require the model to identify the antecedent of an ambiguous pronoun within context, or involve completing a sentence given multiple choices. 
For commonsense knowledge mining, \citet{feldman2019commonsense} construct a
candidate piece of knowledge as a sentence, then use a language model to approximate the likelihood of the text as a proxy for its truthfulness.

Prompts can also be used to explore the linguistic knowledge of PLMs, focusing on different phenomena such as analogies \cite{NEURIPS2020_1457c0d6}, negation \cite{ettinger-2020-bert} or semantic similarity \cite{sun2021ernie}.
Linguistic evaluation of language models \cite{linzen-tacl-2016-assessing,gulordava-etal-2018-colorless,goldberg2019assessing,tran-etal-2018-importance,bacon2019does,mccoy-etal-2020-syntax,linzen-2020-accelerate} usually considers minimal pairs of grammatical and non-grammatical sentences addressing a specific phenomenon that differs in a single place in the sentence. To succeed, a model must score the grammatical sentence higher than its ungrammatical counterpart. A main resource in this context is BLiMP
\citep[Benchmark of Linguistic Minimal Pairs, ][]{warstadt-etal-2020-blimp-benchmark} which provides minimal pairs for various grammatical phenomena. Recently, the use of this benchmark was adapted for language acquisition research \cite{HSFR21}: the authors probe a RoBERTa-based model pre-trained on transcriptions of child-directed speech \cite{macwhinney2000childes} to complete the benchmark task. The preference score can be calculated either {\it holistically}, summing the cross-entropy errors at each position in the sentence \cite{zaczynska-german,HSFR21}, or in an {\it MLM-based} way, where each candidate sentence is  masked by a language model multiple times with the mask changing position. The score
is computed by summing the log-losses at the different masked positions \cite{salazar-etal-2020-masked}.

\paragraph{Other tasks.} 
The PET procedure \cite{schick-schutze-2021-exploiting} was also applied to the Textual Entailment task. 
QA is addressed in ~\citet{khashabi-etal-2020-unifiedqa} with appropriate prompts from the context and questions, formulating several QA tasks into a unified text generation problem with encoder-decoder pre-trained models such as T5. 

Prompts have also been used for the evaluation of text generation. \citet{yuan2021bartscore} used prompts in the BARTSCORE-PROMPT variant of the BARTSCORE measure they propose that treats the evaluation of various text generation tasks as a generation problem. In BARTSCORE-PROMPT, prompts are either appended to the source text or prepended to the target text  and are shown to be useful. For example, adding the phrase “such as” to the translated text when using pre-trained models significantly improves the correlation with human evaluation on German-English machine translation evaluation.

\citet{schick2021selfdiagnosis} showed that PLMs are able to recognize the toxicity of the text they produce (self-diagnosis). Then they proposed an algorithm that permits the language model to produce less problematic text (self-debiasing), using a textual description of the undesired behavior.


\citet{shin2021constrained} explore the use of PLMs as few-shot semantic parsers. The authors use GPT-3 to convert text into a canonical text (in a controlled sub-language) satisfying a grammar, that is then automatically mapped to the target structured meaning representation.

\subsection{Learning from Proxy Tasks} \label{subsec:proxy_tasks}


Templates and prompts play a role again in an indirect approach to NLP tasks called ``proxy tasks''.


Examples for the use of this approach are emotion classification or event and argument extraction, both shown in Figure~\ref{fig:prompt-methods} (right box) with prompt-based proxy tasks. See Table~\ref{tab:prior_work_prompting} for additional examples of proxy tasks and prompt design. 

The key distinction between learning from proxy tasks and previous methods is the use of supervised Natural Language Understanding (NLU) tasks as a proxy instead of self-supervised language modeling for the target task. Indeed, taking advantage of large NLU datasets for extra supervision results in better zero and few-shot performance in the target task with relatively small PLMs \cite{wang2021entailment}, commonly RoBERTa\textsubscript{large} at 345M parameters.  Knowledge-rich classification tasks in particular benefit from PLM proxy tasks, because the latter can reformulate the class label as a prompt, taking advantage of the meaning of class labels instead of treating them as indices.
In this section, we describe the main proxy-task-based learning approaches using QA (Section \ref{subsubsec:extractive_tasks}) and Textual Entailment (Section \ref{subsubsec:classification_tasks}). 

\begin{table*}[]
    \centering
    \scalebox{0.75}{
\begin{tabular}{|p{0.15\linewidth}|p{0.15\linewidth}|p{0.3\linewidth}|p{0.6\linewidth}|}
\hline
\textbf{Application} & \textbf{Work} & \textbf{Task design} & \textbf{Prompt design} \\
\hline
Relation Extraction & \citet{li-etal-2019-entity} & Use question-answering to identify the most appropriate entity span, given an incomplete text and an indication of the class type & Input: The armory is north of the music center. Prompt: Find a facility near $E_1$? $E_1$, physical, facility  \\
& \citet{agirre2021} & Use textual entailment to determine the likelihood of a candidate relation (such as PlaceOfDeath(X,Y) given an input sentence. & Input: Gary's car crash occurred in Houston; Prompt: Gary died in Houston\\
\hline
Event Extraction & \citet{du-cardie-2020-event} & Use a series of ordered questions, each leveraging the output of the previous answer, to find event triggers and appropriate arguments & (1) Input: Donna purchased a new laptop; Prompt: What is the trigger? \underline{purchased}  (2) Prompt: What was purchased? \underline{laptop}   \\
\hline
Topic and Sentiment Classification & \citet{yin-etal-2019-benchmarking} & Use textual entailment to determine whether a topic name $T$ is suitable for a text. & Input: Dinosaurs and humans never coexisted. Prompt: This text is about $T$. \\
& \citet{puri2019zeroshot} & Use question answering to probe for a topic or sentiment name from among a closed set of responses. & Input: Dinosaurs and humans never coexisted. Prompt: How is the text best described? $T_1$, $T_2$, or $T_3$ \\
\hline
Coreference Resolution & \citet{wu-etal-2020-corefqa} & Use question-answering to find a coreferent mention of a marked mention from within the same text. & Input: I arrived at the party with my tux on, and introduced myself as George. I told them that $<$mention$>$ I $<$/mention$>$ was hired to do some Christmas music; Prompt: Who does {it I} refer to? \\
\hline
\end{tabular}}
    \caption{Examples of task design and example prompts for four different applications of prompt-based proxy tasks.}
    \label{tab:prior_work_prompting}
\end{table*}

\subsubsection{Question Answering as Proxy Task} \label{subsubsec:extractive_tasks}

In a strong move away from traditional information extraction, recent studies replace modeling of explicit entity, relation, and event classes with natural language questions that get at the exact item of interest. Questions can be used to probe for the required information in the text.

The choice of using QA as a proxy task is motivated by the relative ease of answering simple questions, as compared to  performing expert annotation for complex linguistic phenomena. 

In information extraction tasks, question prompts typically address identification and classification jointly, by constructing the question to identify a particular type. For example, the question ``Who bought something?" will produce an answer specific to the {\it Buyer} argument role in an event of type Exchange-Ownership (see Figure \ref{fig:prompt-methods}, right box). 

\citet{li-etal-2020-unified} formulates {\bf Named Entity Recognition (NER)} as a QA problem. For example, the prompt ``which person is mentioned in the text?" will identify a mention classified as a PERSON. The proposed BERT-based system performs detection of multiple spans through the use of separate binary classifiers identifying start and end tokens. The authors incorporate synonyms and examples into the queries.

\citet{wu-etal-2020-corefqa} formulated {\bf coreference resolution} as a span prediction task via QA, where a query is generated for each candidate mention using its surrounding context, and a span prediction module uses the query to extract the coreference spans in the document.

\citet{levy-etal-2017-zero} first formulated relation extraction as a QA task. 
This approach has been pursued  in the context of PLMs by \citet{li-etal-2019-entity} and \citet{zhao-2020-questions}.
\citet{han2021ptr} addresses relation extraction with sub-prompts for {\bf entity recognition} and {\bf relation classification}, composing them into a complete prompt using logic rules. Both types of questions are used to probe a QA system in a supervised setting to perform the two sub-tasks. Task decomposition is also used in the work of \citet{zhou2021role} for event extraction where natural questions for {\bf argument identification} (``What plays the role?") and {\bf argument classification} (``What is the role?") mutually improve each other. 

\citet{chen-etal-2020-reading} reformulated {\bf event extraction} as a cloze task with QA model based on BERT and the SQuAD 2.0 dataset \cite{rajpurkar2018know}. Question answering is used directly, preserving the QA format, in \citet{du-cardie-2020-event}, \citet{feng2020probing}, \citet{li-etal-2020-event}, \citet{zhou2021role} and  \citet{liu-etal-2020-event} for argument extraction, including the argument identification and classification sub-tasks. In these cases the event extraction training data is converted to the QA format, where the questions are derived from the ontology. \citet{liu-etal-2020-event} also experimented in a zero-shot setting where no task-specific data is used for training, only using prompts for probing. The zero-shot setting for the full event extraction pipeline has been explored in \citet{lyu-etal-2021-zero} where QA-based prompts are used for argument extraction and prompts based on Textual Entailment \cite{dagan2013recognizing} are used for trigger classification (see Section~\ref{subsubsec:extractive_tasks} below). Several ablation experiments analyzed the different components of the system such as the choice of PLM, the choice of QA dataset and the way to generate the questions (fixed vs. contextualized). It was shown in particular that RoBERTA trained on QAMR \citep{michael-etal-2018-crowdsourcing} achieved the best results for argument extraction. 

Identification-only sub-tasks such as {\bf trigger identification} \cite{du-cardie-2020-event}, are addressed by more general questions, e.g. ``What is the trigger?''. In contrast,  \citet{zhou2021role} uses separate questions to address the identification and classification of arguments. 

\citet{du-etal-2021-qa} addressed {\bf slot-filling}, which aims to extract task-specific
slot fillers (for example, a flight date) from user
utterances 
by formulating it as a QA task. In particular, they addressed the zero-shot slot-filling problem, where the model needs to predict spans and their values, given utterances from new, unsupervised domains. Extracting slot-filler spans from utterances with a QA model improved the performance, compared to a direct encoding of the slot descriptions.

Lastly, \citet{gao-etal-2019-dialog} formulated the {\bf dialogue state tracking} task that aims to the estimate of the current belief state of a dialog given all the preceding conversation, as a QA problem. The proposed system uses a simple attention-based neural network to point to the slot values within the conversation. This direction was pursued by \citet{gao-etal-2020-machine} who also included a multiple-choice setting, where several candidate values for each slot in the question are given. The latter setting was also investigated by \citet{zhou2020multidomain} who further improved the results. \citet{namazifar2020language} used this approach to address language understanding problems in the dialogue context, experimenting on ATIS (Airline Travel Information Systems, \citealp{hemphill-etal-1990-atis}) and on the Restaurants-8k dataset \cite{coope-etal-2020-span}.

\paragraph{QA Task Design.}
Questions are typically generated via hand-crafted templates derived from the task-specific ontologies. Some of the works introduce contextualization, integrating relevant words from the text into the question. For example, in argument extraction, the question can include the trigger extracted from the text \cite[e.g][]{liu-etal-2020-event,lyu-etal-2021-zero} or another argument that was previously identified \cite{li-etal-2020-event} (see the Event Extraction row in Table~\ref{tab:prior_work_prompting}). Neural based question generation models can also improve the quality of the question, as in \citet{liu-etal-2020-event}, where monolingual unsupervised machine translation \cite{lample-etal-2018-phrase} is used to generate the part of the question that does not depend on the template, translating a descriptive statement into a question-style expression.

Other aspects of QA-style proxy tasks are the ability to use multiple questions, and to formulate questions in any style. In addition to sequential questions for determining event arguments, multiple formulations of the same question may be used in a weighted voting scheme to generate an ensemble answer \citet{zhao-2020-questions}. The input to the QA system need not necessarily include natural questions. It may instead consist of pseudo-questions such as keywords, synonyms, position index of labels, or a single word/type from the ontology or annotation guidelines \cite[e.g.][]{li-etal-2020-unified, du-cardie-2020-event}.

PLMs fine-tuned on the SQuAD 2.0 dataset \cite{rajpurkar2018know} or on QAMR are particularly useful to initialize QA-style prompt-based learning methods.\footnote{Fine-tuning on a PLM on QAMR corresponds to the p-QuASE representation presented in \citet{he-etal-2020-quase}.} With the advent of web-scale QA datasets~\cite{huber2021ccqa}, QA-infused PLMs may provide significantly richer representation, enabling a wider range of applications.





\subsubsection{Textual Entailment as Proxy Task} \label{subsubsec:classification_tasks}

Textual Entailment is a popular proxy for classification tasks \cite{yin-etal-2019-benchmarking}, as these models have shown a striking ability to perform few-shot learning.  \citet{wang2021entailment} hypothesizes that this phenomenon might be because the entailment task is a true language understanding task; a model that performs entailment well is likely to succeed on similarly-framed tasks. An example of textual entailment as a proxy for {\bf emotion classification} is shown in Figure~\ref{fig:prompt-methods}, while an example of its use for {\bf topic detection} is shown in Table~\ref{tab:prior_work_prompting}.

For entailment prompting, developers define a template that describes the task, and create a natural language version (``verbalization'') of each potential label. Multiple hypotheses for entailment are produced by inserting the potential labels into the template. The inference is performed by selecting the most probable candidate hypothesis given the input. Some recent works also make use of multiple verbalizations for each label to boost the system performance \cite{sainz-rigau-2021-ask2transformers, agirre2021}.

\citet{agirre2021} also proposed an approach to guiding the ``art'' that is prompt crafting more towards a ``science'': the authors fine-tune a model on Textual Entailment data and use the model's probability of a prompt given the template, applied on the guideline example(s), to measure the quality of manually designed prompts. 

\citet{obamuyide-vlachos-2018-zero} reformulated {\bf relation extraction} as a textual entailment task. This approach has been pursued in the context of PLMs  by \citet{agirre2021}.

Roughly equivalent to textual entailment is Yes/No Question Answering \cite{clark-etal-2019-boolq} where a model is asked about the veracity of some fact given a passage. It has also been used as a proxy task for text classification by \citet{zhong2021adapting}.


PLMs needs to be fine-tuned to solve the textual entailment task. They are commonly fine-tuned on MNLI \cite{williams-etal-2018-broad}, but other datasets such as SNLI \cite{bowman-etal-2015-large}, FEVER \cite{thorne-etal-2018-fever}, ANLI \cite{nie-etal-2020-adversarial} or XNLI \cite{conneau-etal-2018-xnli} are also used. In addition, data from different tasks can be used when framed properly \cite{zhong2021adapting}.

\begin{figure*}[h!]
  \centering 
  \includegraphics[scale=0.4]{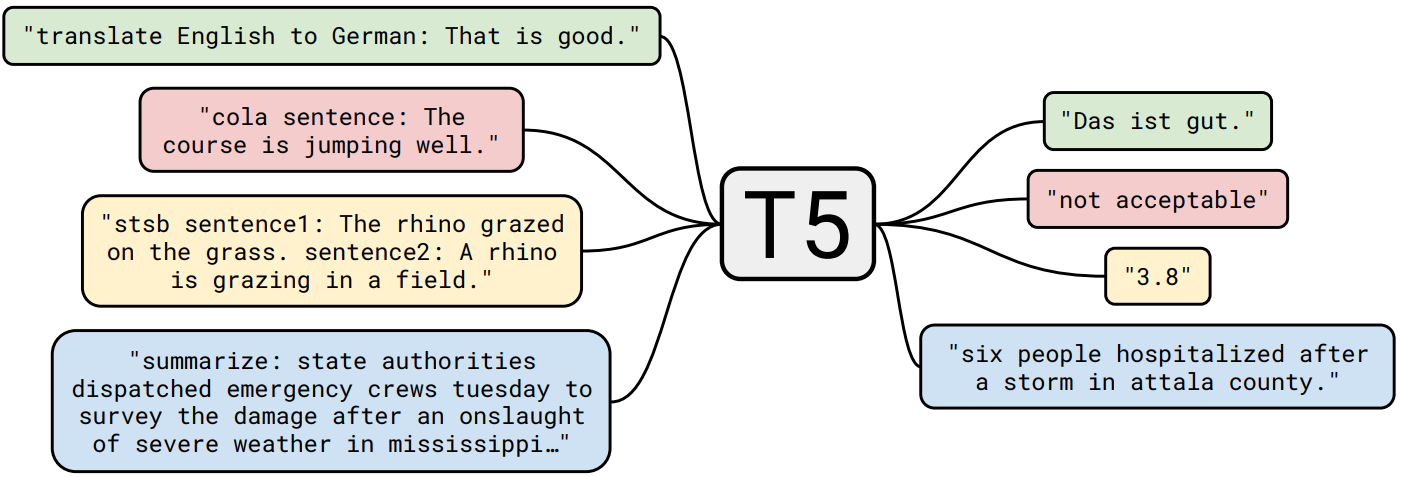}
  \caption{An illustration of T5~\cite{raffel2020exploring} text-to-text text generation approach for Machine Translation, linguistic acceptability, text semantic similarity and summarizing tasks. Figure source: ~\citet{raffel2020exploring}.}
  \label{fig:t5}
\end{figure*}

\section{Paradigm 3: NLP as Text Generation} \label{sec:paradigm3}






The success of generative Transformer-based PLMs\footnote{In this section and next, we use the term PLM to refer to a generative PLM. } such as GPT, BART, and T5 has recently sparked interest in leveraging generative PLMs to solve various non-generative NLP tasks. These tasks include, but are not limited to, traditional discriminative tasks such as classification and structure prediction. For example, Figure~\ref{fig:t5} illustrates this ``text-to-text'' approach as described in ~\citet{raffel2020exploring}. Instead of using traditional discriminative models for NLP tasks, these tasks are reformulated as text generation problems so that they can be directly solved with generative PLMs. The generated output sequences usually include the desired labels or other auxiliary information for the given task, enabling accurate reconstruction of the expected class labels (i.e. to avoid ambiguities in mapping) and facilitating the generation/decoding process (i.e. to provide sufficient context for predictions). 

It is worth noting that some NLP tasks are already text generation tasks. Therefore, a straightforward strategy for those tasks is to fine-tune a generative PLM using task-specific training data to perform the specific tasks of interest. Examples include Machine Translation~\cite{cooper-stickland-etal-2021-recipes}, text summarization~\cite{lewis-etal-2020-bart}, text style transfer~\cite{lai2021thank}, etc. We refer readers to Section~\ref{sec:paradigm1} for more detailed discussion of this ``pre-train then fine-tune'' approach. In this section, we focus on tasks that are not traditionally text generation tasks.

\paragraph{Reformulating NLP Tasks as Text Generation Problems}

Pre-trained from large corpora, PLMs demonstrate an extraordinary ability to generate text. PLMs also capture rich knowledge that could be used for many NLP tasks and show strong performance on learning new patterns via fine-tuning. These factors lead to the hypothesis that many NLP tasks can be reformulated as text generation problems. In particular, given an NLP task with an input text $x$, this approach first attempts to design an output sequence $y$ that includes information about the desired labels for $x$ (e.g. markers). Then, a PLM directly generates $y$, conditioning on the input $x$, modeling $P(y|x)$. In this formulation, the desired labels/outputs for the task on $x$ need to be retrieved unambiguously from $y$, requiring $y$ to be generated in a valid format by the design of the reformulated task. In addition to the label information, evidence useful for providing context can also be incorporated into the formulation of $y$ to aid the generation process. To train the PLMs, the original training data of the NLP task is first converted into pairs $(x,y)$ following the designed format. The PLMs are usually fine-tuned with such pairs using the standard maximum likelihood loss. 

There are a few advantages of this approach. First, in this formulation, a unified text-to-text/seq2seq framework can be used to solve different NLP tasks via encoder-decoder architectures, thus facilitating multi-task learning and transfer learning across tasks of different natures \cite{raffel2020exploring}. Second, the direct generation of labels in output sequences allows the PLMs to exploit the semantics of the labels to improve the performance and data efficiency, a benefit that cannot be achieved in discriminative models \cite{Paolini2021Structured}. Finally, when adapting to structure prediction problems, PLM-based models can naturally capture the inter-dependencies between prediction steps/tasks in the modeling process to further improve the performance \cite{athiwaratkun2020augmented}.

As such, the formation of the output sequence $y$ for an input $x$ is critical for the performance of the PLM-based methods. Existing works tend to customize such output sequences for specific NLP tasks to better capture the nature of the tasks. Therefore, in the rest of this section, we group prior works according to their strategies in designing the output sequences to solve NLP tasks with generative models, and discuss their representative techniques in each subsection. Table~\ref{tab:NLPbyGTM} provides a brief summary. 


\begin{table*}[]
    \centering
    \scalebox{0.49}{
    \begin{tabular}{|p{0.1\linewidth}|p{0.2\linewidth}|p{0.37\linewidth}|p{0.6\linewidth}|p{0.65\linewidth}|}
        \hline
        \multirow{2}{1cm}{\textbf{Output Type}} & \multirow{2}{*}{\textbf{Work}} & \multirow{2}{*}{\textbf{Task}} & \multicolumn{2}{c|}{\textbf{Example}} \\ \cline{4-5}
        & & & \textbf{Input} & \textbf{Output} \\ \hline
        \multirow{19}{1cm}{Label-augmented Text} & \multirow{12}{*}{\citet{Paolini2021Structured}} & \multirow{2}{*}{Joint Entity and Relation Extraction} & Tolkien’s epic novel The Lord of the & [ Tolkien \textbar{} person ]‘s epic novel [ The Lord of the Rings \\
        & & & Rings was published in
1954-1955. & \textbar{} book \textbar{}
author = Tolkien ] was published in 1954-1955 \\ \cline{3-5}
        & & \multirow{3}{*}{Relation Classification} &  Born in Bologna, Orlandi was a student of the famous Italian & relationship between [ Carmen Melis ] and \\
        & & & [ soprano ] and voice teacher [
Carmen Melis ] in Milan. & [ soprano ] = voice type \\
        & & & The relationship between [ Carmen Melis ] and [ soprano ] is &  \\ \cline{3-5}
        & & \multirow{2}{*}{Semantic Role Labeling} & The luxury auto maker last year [ sold ] & [ The luxury auto maker \textbar{} subject ] [ last year \textbar{} temporal ]  \\
        & & & 1,214 cars in the U.S. & sold [ 1,214 cars \textbar{} object ] [ in the U.S.
\textbar{} location ] \\ \cline{3-5}
        & & Event Extraction & Two soldiers were attacked and injured yesterday & Two soldiers were [ attacked \textbar{} attack ] and [ injured \textbar{} injury ] yesterday \\ \cline{3-5}
        & & \multirow{2}{*}{Coreference Resolution} & Barack Obama nominated Hillary Rodham Clinton & [ Barack Obama ] nominated [ Hillary Rodham Clinton ] as [ his \textbar{} \\
        & & & as his secretary of state on Monday. & Barack Obama ] [
secretary of state \textbar{} Hillary Rodham Clinton ] on Monday \\ \cline{3-5}
        & & \multirow{2}{*}{Dialogue State Tracking} & [ user ] : I am looking for a cheap place & [ belief ] hotel price range cheap, hotel \\
        & & & to stay [ agent ] : How long? [ user ] : Two & type hotel, duration two [ belief ] \\ \cline{2-5}
        & \multirow{2}{3cm}{\citet{athiwaratkun2020augmented}} & Slot Filling & Add Kent James to the Disney soundtrack & (( AddToPlaylist )) Add [ Kent James \textbar{} artist ] to the [ Disney \textbar{} playlist ] \\ \cline{3-5}
        & & Named Entity Recognition & He is John Wethy from NBC News & He is [ John Wethy \textbar{} person ] from [ NBC News \textbar{} org ] \\ \cline{2-5}
        & \multirow{5}{*}{\citet{Zhang2021Towards}} & \multirow{2}{*}{Aspect Opinion Pair Extraction}  & \multirow{2}{*}{Salads were fantastic, our server was
also very helpful.} & [Salads
\textbar{} fantastic] were fantastic, \\
        & & & & our [server \textbar{}
helpful] was also very helpful. \\ \cline{3-5}
        & & \multirow{2}{*}{Aspect Sentiment Triplet Extraction} & \multirow{2}{*}{The Unibody construction is solid,
sleek and beautiful.} &  The
[Unibody construction \textbar{} positive \textbar{} solid, sleek, \\
        & & & & 
beautiful] is solid, sleek and beautiful. \\ \cline{3-5}
        & & Target Aspect Sentiment Detection & The
pizza was cold. & The [pizza \textbar{} food
quality \textbar{} negative] was cold. \\ \hline
        \multirow{6}{1cm}{Generating Word Indices} & \citet{yan2021unified} & Named Entity Recognition & have muscle pain and fatigue & 2 3 7 2 5 6 \\ \cline{2-5}
        & \multirow{4}{*}{\citet{yan2021unifiedb}} & Aspect Term Extraction & \multirow{4}{5cm}{The wine list is interesting and has good values , but the service is dreadful} & 1, 2, 12, 12 \\ \cline{5-5}
        & & Opinion Term Extraction & & 4, 4, 7, 8, 14, 14 \\ \cline{5-5}
        & & Aspect-level Sentiment Classification & & 1, 2 , Positive \\ \cline{5-5}
        & & Aspect-oriented Opinion Extraction & & 1, 2, 4, 4, 7, 8 \\ \cline{2-5}
        & \multirow{2}{*}{\citet{rongali20dont}} & \multirow{2}{*}{Slot Filling} & \multirow{2}{*}{play the song don't stop believin by journey} & PlaySongIntent SongName( @pt r3 @pt r4 @pt r5) \\
        & & & & SongName ArtistName( @pt r7 )ArtistName \\ \hline
        \multirow{2}{1cm}{Generating Answers} & \citet{wang-etal-2021-generative} & Closed-book QA & What is Southern California often abbreviated as? & Southern California, often
abbreviated SoCal, is $\ldots$ ANSWER SoCal \\ \cline{2-5}
        & \citet{hsu2021answer} & Answer Selection & How a water pump works? & A water pump is a device that moves fluids by mechanical action. \\ \hline
        \multirow{4}{1cm}{Filling Templates} & \multirow{2}{*}{\citet{du2021template}} & \multirow{2}{*}{Event Extraction} & [CLS] Attack, Bombing, Arson, $\ldots$ [SEP\_T] (Document & \multirow{2}{*}{[CLS] Attack -T1 REEs- [SEP\_T] Bombing -T2 REEs- [SEP\_T]} \\
        & & & tokens): Several attacks were carried out in La Paz $\ldots$ [SEP] & \\ \cline{2-5}
        & \multirow{2}{*}{\citet{li-etal-2021-document}} & \multirow{2}{*}{Event Argument Extraction} & Elliott testified that on April 15, McVeigh came into the body -tgr- reserved -tgr- the truck, to be picked up at
     4pm two days later
shop and & Elliott bought, sold or traded truck to McVeigh in exchange for \$280.32 for the benefit of -arg- at body shop place \\
        \hline
        \multirow{3}{1cm}{Structure-linearized Texts} & \citet{ren2021hyspa} & Joint Entity and Relation Extraction & He was captured in Baghdad late Monday night & ``He" Type PER [SEP] ``Baghdad" Type GPE PHYS ``He" \\ \cline{2-5}
        & \multirow{2}{*}{\citet{lu2021text2event}} & \multirow{2}{*}{Event Extraction} & \multirow{2}{*}{The man returned to Los Angeles from Mexico} & ((Transport returned (Artifact The man) (Destination Los Angeles) \\ 
        & & & & (Origin Mexico)) \\ \hline 
        \multirow{6}{1cm}{Ranking Input-output Pairs} & \multirow{2}{3cm}{\citet{nogueira2020beyond}} & \multirow{2}{*}{Answer Selection} & \textless bos\textgreater Ice formations in the Titlis glacier cave & \multirow{2}{*}{0.5} \\
        & & & \textless boq\textgreater How are glacier cave formed \textless eoq\textgreater & \\ \cline{2-5}
        & \multirow{2}{*}{\citet{nogueira2020document}} & \multirow{2}{*}{Document Retrieval} & How are glacier cave formed [Q] A glacier & \multirow{2}{*}{True} \\
        & & & cave is a cave formed within the ice of a glacier [D] & \\ \cline{2-5}
        & \citet{Cao2021AutoregressiveER} & Entity Retrieval & Superman saved [START] Metropolis [END] & Metropolis (comics) \textbar{} Metropolis (1927 film) \\ \cline{2-5}
        & \citet{cui2021template} & Named Entity Recognition & ACL will be held in Bangkok & Bangkok is a location \\ \hline
    \end{tabular}}
    \caption{A summary of methods reformulating NLP task as a generation task solved by PLMs. }
    \label{tab:NLPbyGTM}
\end{table*}

\subsection{Generating Label-Augmented Texts}~\label{sec:label-aug-text}

In this strategy, the output sequence $y$ copies the input text $x$ and augments it with additional markers that can be decoded into desired label annotations for $x$ for a given NLP task. The repetition of the words from the input text aims to provide explicit context to reduce ambiguity for the generation process \citep{Paolini2021Structured}. This strategy is often applied to structure prediction tasks that aim to jointly extract the text spans of interest and their relations or dependencies in an input text. 

\citet{athiwaratkun2020augmented} explores the idea of label-augmented text generation for sequence labeling problems, e.g. slot filling (identifying spans that define the left or right ``slot'' of a relationship) and Named Entity Recognition (NER). Given an input sentence, the output sequence is formed by marking the token sequences for the slots or entity types of interest, for instance with square brackets or another identifier. The corresponding labels are then introduced immediately after the token sequences, within the brackets, separated by the token by a bar ``$|$''. 
The encoder-decoder PLM T5 is used to generate label-augmented texts. \citet{Paolini2021Structured} extends this idea to other structure prediction tasks, including joint entity and relation extraction, relation classification, semantic role labeling (SRL), event extraction, coreference resolution, and dialogue state tracking. To encode a relation between two text spans the input text, the second text span might be annotated with both the relation label and an indicator of the first text span. 

For example, for the joint entity and relation extraction task, the input sentence $x$ can be transformed into the label-augmented output sequence $y$, where (1) the square brackets indicate token spans for entity mentions; (2) \texttt{person} and \texttt{book} are the corresponding entity type labels; and (3) \texttt{author=Tolkien} indicates the author relation between \texttt{Tolkien} and \texttt{The Lord of the Rings}:

\begin{mdframed}

\hspace{3mm} $x= \texttt{Tolkien’s epic novel The }$
\texttt{Lord of the Rings was published in 1954-1955.}

\vspace{2mm}

$y= \texttt{[Tolkien|person]’s epic }$ 
\texttt{novel [The Lord of the Rings|book|author=Tolkien] was published in 1954-1955.}

\end{mdframed}

In order to transform the generated label-augmented texts into desired annotations, \citet{Paolini2021Structured} uses dynamic programming to match the generated output sequence and the input text, searching for the closest entity mention that exactly matches the predicted tail entity and discarding invalid entity/relation types. Similarly, \citet{Zhang2021Towards} utilize label-augmented text generation for different variations of aspect-based sentiment analysis (ABSA), including aspect opinion pair extraction, unified ABSA, aspect sentiment triplet extraction, and target aspect sentiment detection. \citet{Zhang2021Towards}, also propose a normalization prediction mechanism: if a generated token does not belong to the original sentence or set of expected labels, the closest word from the input sentence using the Levenshtein distance is used instead.

Due to the unified text-to-text formulation, label-augmented text generation allows multi-task learning where a single generative model can be trained to simultaneously perform multiple tasks of different natures. \citet{Paolini2021Structured} and \citet{athiwaratkun2020augmented} show that learning from multiple tasks with a single model can improve the performance on the individual tasks. Furthermore, label-augmented text generation also shows impressive performance in few-shot learning settings \cite{Paolini2021Structured}, improving the data efficiency. 

\subsection{Generating Word Indices}

For many text understanding problems (e.g. span tagging problems such as NER), the generative PLM must not generate words that are not in the input text, other than markers or labels as shown in the example in Section~\ref{sec:label-aug-text}. Restricting the PLMs to consider only words in the input text as candidates at decoding (text generation) time enforces this constraint.

An alternative approach is to directly generate {\it indices} of the words of interest in the input text. Given the input $x$, the output sequence $y$ provides a sequence of index numbers referring to the positions of words in $x$. Label indices encode class labels within $y$. A few examples are included in Table~\ref{tab:NLPbyGTM} in the ``Generating Word Indices'' rows.

\citet{yan2021unified} explores an index generation idea for NER that can naturally handle different settings, e.g. flat, nested, and discontinuous NER. Given the input sequence $x = [x_1,x_2,\ldots,x_n]$, the output sequence $y$ is formed via the indices: $y = [s_{11},e_{11},\ldots,s_{1_{k_1}},e_{1_{k_1}},t_1,\ldots, s_{i1},e_{i1},\ldots,$
$s_{i_{k_i}},e_{i_{k_i}},t_i]$ where $s$ and $e$ indicates the start and end indexes of a span. The spans for the $i$-th name in $x$ are represented by the tuple $[s_{i1},e_{i1},\ldots,s_{i_{k_i}},e_{i_{k_i}},t_i$] where $t_i$ is the index of the entity type and $k_i$ is the number of text spans for the $i$-th name (a name can have multiple spans due to the consideration of discontinuous names). As such, $s_{i_j}$ and $e_{i_j}$ should be between 1 and $n$ while the entity types can be indexed from $n+1$ (i.e., $t_i > n$). To compute the hidden vectors at  decoding time, the representations for the span indices can be obtained from the representations of the corresponding words in the input sentence $x$ (i.e., via pointer networks~\citep{vinyals2015pointer}). BART is used as the base model for the index generation for NER. 

Similarly, \citet{yan2021unifiedb} generates indices of the spans of interest for variations of the aspect-based sentiment analysis (ABSA) task, including aspect term extraction, opinion term extraction, aspect-level sentiment classification and aspect-oriented opinion extraction. Finally, casting a problem into an index generation task is also proposed for semantic parsing (i.e. filling slots) \cite{rongali20dont}. The output sequence in this work starts with the intent, followed by slot names and the index sequences of the words in the input for the slots. At decoding time, each step produces a distribution over the word indexes in the input sentence (as a pointer network) and the vocabulary for slots and intents in the datasets.




\subsection{Generating Answers}

This strategy is designed mainly for the QA task. The basic idea is to fine-tune PLMs to generate answers for the QA problems of interest. \citet{wang-etal-2021-generative} use BART for closed-book QA that aims to directly provide answers for input questions. They show that BART struggles on a version of SQuAD for closed-book QA where the test and training data do not have much question and answer overlap. It also shows that BART cannot remember knowledge from the fine-tuning data if there are many training passages for fine-tuning. Suggestions to address those issues include decoupling the knowledge memorization and QA fine-tuning, and forcing the model to recall relevant knowledge in the answer generation step. 

\citet{hsu2021answer} applies answer generation to the problem of answer selection, in which the system must choose the correct answer from a provided candidate set (it is also provided the question). Instead of training an answer {\it selector} \citep{han2021modeling}, \citet{hsu2021answer} uses answer {\it generation} through fine-tuning PLMs such as T5 and BART, which consume the input question and the top answer candidates, then generate an answer for the question. To prepare training data for  fine-tuning, the output answers might come from human annotators or be directly inherited from the provided correct answer (i.e. the correct answer will be removed from the input for the generative models and maybe replaced by another answer candidate).


\subsection{Filling templates}

For many extraction tasks, the output are spans organized into one or several templates. For example, event extraction tasks require a system to extract templates in the form of {\it who did what to whom where and when}. 

A template defines the appropriate relationship and order for the spans and labels for generation, forming the output sequence $y$. \citet{du2021template} explores the template filling idea for an IE task: given a document, a model must identify event templates/types (via trigger words) and entity mention fillers for the argument roles. A sequence-to-sequence model for template filling takes the possible event types concatenated with words in the input document $x$ as the input, and outputs a sequence of tuples. Each tuple corresponds to a detected event template, starting with an event type and followed by the text span fillers for the roles in the input document (following an order). The roles with no fillers are associated with {\it null}. \cite{Zhang2021Towards} also examines a similar approach of tuple generation for ABSA.

The template filling methods can also introduce additional information into the templates to aid the label generation process, such as natural descriptions or definitions of the labels. In particular, \citet{li-etal-2021-document} pursue a general template filling approach for document-level event argument extraction: given an event trigger in an input document, find entity mentions to fill in the roles for the event. A conditional generative model (e.g. BART) is employed for argument extraction where the input (the condition) to the model is created by combining an unfilled template and the document context. The template is essentially a sentence describing the event type augmented with placeholders for argument role fillers. The output sequence $y$ is a filled template where placeholders are replaced by concrete arguments (entity mentions). To avoid entity type mismatch for arguments, the templates in the inputs are also appended with sentences to indicate entity types for arguments (e.g. $arg_1$ is a person) that can be used to re-rank the output sequences to follow the type constraints. Below is an example input $x$ in which a template over a list of event arguments $arg_1,...,arg_6$ and the document text DOC\_TEXT are concatenated, and output $y$, in which the underlined text spans are fillers from DOC\_TEXT) from \citet{li-etal-2021-document}:

\begin{mdframed}

\hspace{2mm} \texttt{$x$ = $\langle s \rangle$ $\langle arg_1 \rangle$ bought, sold, } \\ 
\hspace{0mm} \texttt{or traded $\langle arg_3 \rangle$ to $\langle arg_2 \rangle$ in exchange for $\langle arg_4 \rangle$ for the benefit of $\langle arg_5 \rangle$ at $\langle arg_6 \rangle$ place. $\langle s \rangle$ $\langle /s \rangle$ DOC\_TEXT $\langle /s \rangle$}

\vspace{2mm}

\texttt{$y$ = \underline{Elliott} bought, sold or traded \underline{truck} to \underline{McVeigh} in exchange for \underline{$280.32$} for the benefit of $\langle arg \rangle$ at \underline{body shop} place.}

\end{mdframed}


\subsection{Generating Structure-Linearized Texts}


Structure prediction problems in NLP typically require multiple prediction outputs for an input text $x$ that are interconnected to form a single structure that represents the input. To cast structure prediction tasks as text generation problems, one approach involves linearizing the output structure to serve as the output sequence $y$. For example, taking $x$ as input, TEXT2EVENT~\cite{lu2021text2event} directly generates the event structures $y$:

\begin{mdframed}

\hspace{2mm} $x=$ \texttt{The man returned to Los Angeles from Mexico following his capture Tuesday by bounty hunters.} 

\vspace{2mm}

$y=$ \texttt{((Transport returned (Artifact The man) (Destination Los Angeles) (Origin Mexico)) (Arrest-Jail capture (Person The man) (Time Tuesday) (Agent bounty hunters))}

\end{mdframed}

Graph traversal algorithms are often used to accomplish the linearization in this approach. \citet{ren2021hyspa} study structure-linearization for joint entity and relation extraction \citep{li2014constructing,miwa2016end}. The main idea is to construct an information graph for each input sentence to capture entity mentions, their entity types, and relations. Depth or breath first traversal can be used for graph linearization for $y$. To solve the sequence-to-sequence problem for pairs of $\langle x,y \rangle$, \citet{ren2021hyspa} linearize the information graph to an alternating sequence of nodes and edge types (given depth/breath first traversal), and directly generate such sequences via a hybrid span decoder that decodes both the spans and the types recurrently. For event extraction with joint extraction of event triggers and arguments \citep{li2014constructing,nguyen2016jointevent}, a structure-linearization and text generation approach comes from \citet{lu2021text2event}.  The authors first build a labeled tree to capture the event types and argument roles in the sentence (i.e. event schema), with trigger and argument text spans as leaves. The labeled tree is transformed into the output sequence $y$ by depth-first traversal where T5 is used to perform the conditional generation of $y$ from $x$. To improve the model, a trie-based constrained decoding procedure \citep{chen2020parallelsentence,Cao2021AutoregressiveER} is introduced to ensure the generation of valid event structures. A trie (prefix-tree) determines possible candidates for the next generation step given the previously generated tokens to guarantee valid output sequences. \citet{lu2021text2event} also report the effectiveness of the generation-based model for extending models to extract new event types.


\subsection{Ranking Input-Output Pairs}

Some NLP tasks require choosing the best response from among many: answer selection in multiple choice-sytle QA, information retrieval, and certain kinds of entity retrieval all provide a set of candidate answers to a posed query from which the system selects the best one. Typically, a system will rank the candidates in relation to the input query, a task at which PLMs can excel. The idea has its roots in the classical literature on probabilistic models for information retrieval that rank documents using language models \cite{ponte1998language,lafferty2001document}. Given an input query, a candidate document is scored in two steps: (i) training a language model on the candidate document, and (ii) computing the likelihood of generating the input query from that language model, which  serves as the candidate's ranking score. 

We now see the use of PLMs to perform  generation-based ranking for selection. \citet{nogueira2020beyond} apply the idea for answer selection by fine-tuning generative models (GPT-2 or BART) over $\langle$answer, question$\rangle$ pairs, thus learning to generate questions given correct answer passages. The simplest approach is to fine-tune the models over only the positive pairs. \citet{nogueira2020beyond} also explore fine-tuning with negative pairs using an unlikelihood objective or ranking-based objective (e.g. the hinge loss). At inference time, the ranking score for an input passage is obtained via the likelihood of the fine-tuned PLM over the input question conditioning on that passage. 

\citet{nogueira2020document} approach the document relevance ranking problem in a similar way. The paper concatenates the input query and each candidate document and feeds them as an input/condition for a fine-tuned T5 model. To fine-tune T5, the model is asked to generate ``True'' or ``False'' as the output sequence, indicating the document's relevance to the query. The probability of generating ``True'' is used as the ranking score for the candidate. 

\citet{Cao2021AutoregressiveER} address the entity retrieval problem: given a set of Wikipedia articles representing entities, return the entity that is most relevant to a textual input source $x$. Each entity is represented by its textual representation (e.g. the title of its Wikipedia article), which will be used as the output sequence $y$ for the generative models.
BART is fine-tuned to rank the entities using the generation likelihood $P(y|x)$. 
\citet{cui2021template} explore generation-based ranking for NER, especially in few-shot and cross-domain few-shot settings. Given an input sentence and a text span, a template is formed by concatenating the words in the span and an expression of type ``is a {\it entity\_type} entity''. The original sentence and the template serve as an input-output pair in sequence-to-sequence models. BART is then employed to score this pair (using the probability of the template output produced by the decoder of BART). For each span, the entity type corresponding to the template with highest score is selected. Original NER training data is used to create gold standard templates to fine-tune BART.

In addition to question answering, other generative tasks have been shown to benefit from PLMs. For instance, semantic parsing, generating a structure representing the semantics of the sentence, is explored in a recent work by \citet{shin2021constrained}. Authors show that by reformulating the output of PLMs the generated natural language can be used to recover the semantic structure of the input text. They use GPT-3 in the experiments. 

\section{Data Generation via PLM} \label{sec:data_gen}

In addition to using PLMs to perform NLP tasks directly, PLMs can be used to generate data that can be used to enhance the performance of NLP systems in two ways. 
Note that these data generation approaches are complementary to the three paradigms of PLM-for-NLP discussed in previous sections.

First, data generated by PLMs can be combined with original training data to improve NLP models where training data is too sparse. Typically, this is applied to create new labeled data to increase diversity, enrich the models, and otherwise alleviate common limitations of hand-labeled data. 
The studies presented below discuss, for various downstream NLP tasks: approaches for fine-tuning PLMs to ensure they capture the key characteristics of the task when performing data generation;  appropriate reformulation of the original training data for PLM fine-tuning and generation; and filtering the new data for noise introduced by the generation process. 

Second, we discuss the use of auxiliary data generated by PLMs to shed light on interesting aspects of NLP models. This approach plays a role in machine learning explainability by providing generations such as counterexamples, clarifying questions, context for answers, inference rules, and other insight-rich sequences.

\subsection{Augmenting NLP Models with Automatically Generated Data}

Traditional approaches to data augmentation, including generation via semi-supervised learning on large unlabeled data sets and synthesis with back-translation or synonymous word replacement \cite{feng2021survey,chen2021empirical} were shown to be effective for increasing NLP models' accuracy and/or coverage. Newer studies show that PLMs can be also used as an effective method for data augmentation \cite{zhang2020data,yang2020gdaug,peng2020data,kumar2020data,anabytavor2020do}, requiring no significant change to the model architecture. The fluency of PLM text generations stand in contrast to the outcomes of traditional approaches that may produce less natural samples. As discussed in previous sections, the massive amount of linguistic knowledge accumulated by the PLM allows for adaptation to many domains and tasks, including those with very limited labeled data. The vast knowledge may also produce a greater variety of new examples, further improving the NLP models trained on them. 
We organize the discussion of data augmentation methods according to the NLP tasks they support.



\subsubsection{Information Extraction (IE)} Prior works explored synthetic data generation with PLMs \cite{madaan2020eigen,bosselut2019comet} for a variety of IE tasks.

\citet{pouranbenveyseh2021unleash} and \citet{pouranbenveyseh2021augmenting} use GPT-2 to produce synthetic labeled data for event detection. Sentences in existing training datasets are augmented with markers to indicate positions of event trigger words. The resulting labeled sentences are  used to fine-tune GPT-2 using the standard autoregressive next word prediction (NWP) objective. \citet{pouranbenveyseh2021unleash} shows that the fine-tuned GPT-2 model can generate label-augmented data for different domains (e.g. newswire, cybersecurity); however, the generated data might include some noise, for instance, incorrect grammar, meaningless sentences, or incorrect annotations. To minimize the impact of the noisy generated examples and maximize the benefits of the generated data, \citet{pouranbenveyseh2021unleash} and \citet{pouranbenveyseh2021augmenting} present a student-teacher network framework: the teacher network is trained on the original labeled data to obtain anchor knowledge, while the student is trained over the combination of original and synthetic data, with constraints introduced to enforce consistency with the teacher's learned anchor knowledge. The framework leads to significant performance improvement over different datasets for event detection. 

\citet{guoroth2021constrained} employ GPT-2 to generate synthetic labeled data for cross-lingual NER following the annotation projection approach: training data in a source language is translated and projected into a target language to train models. To project annotation, a training sentence in the source language is first translated into the target language using word-to-word translation (via a dictionary). GPT-2 is then fine-tuned to generate complete sentences from the important words in target languages. A hard-constrained generation mechanism is also encoded into the decoding process of GPT-2 to ensure the appearance of the named entities in the original source sentence in the automatically generated sentences.

Synthetic data generation with GPT-2 is also explored for relation extraction in \citet{papanikolaou2020dare}. This paper fine-tunes GPT-2 over labeled examples of the same relation type, where each sentence in the training data is marked with the two entity mentions in the corresponding relation. The fine-tuned model for each relation type is then leveraged to produce new training instances for that relation.


    
\subsubsection{Question Answering (QA)}

Given an input paragraph $C$ and a sampled extractive short answer $A$ in $C$, \citet{alberti2019synthetic} attempts to generate a question $Q$  using a sequence-to-sequence Transformer (with BERT as its encoder). The triple, consisting of the input paragraph, the generated question, and the sampled answer $(C,Q,A)$, can be used as a new training instance for QA models. To mitigate the noise in the generated data, \citet{alberti2019synthetic} present a round trip consistency approach where a second generative model is trained to take the input passage $C$ and generated question $Q$ from the prior step to produce an answer $A'$. The tuple $(C,Q,A)$ is only retained as new training data if  $A' == A$. 

Following a similar principle, \citet{shakeri2020end} explore synthetic data generation for cross-domain QA where models trained on a source domain (typically SQuAD) are evaluated on datasets from a different target domain. The paper aims to generate QA pairs in the target domain and combine them with the source-domain training data to train improved QA models. The data generation model is also trained on the source domain dataset SQuAD using BART and GPT-2. Starting with a passage as the context, the generative models directly generate QA pairs. Generated QA pairs are filtered by the likelihood scores of the generative models to reduce noise.

The data generation idea is extended to multi-hop QA that requires combining disjoint pieces of evidence to answer a question. In particular, \citet{liangming2021unsupervised} aim to generate human-like multi-hop question–answer pairs to train QA models. The model consists of three components: operators, reasoning graphs, and question filtration. Operators are atomic operations that are implemented by rules or off-the-shelf pretrained models to retrieve, generate, or fuse relevant information from input contexts. 
Approaches to fusing relevant information from across contexts include: fine-tuning a T5 model on SQuAD to generate single-hop questions; generating descriptions of table entities with GPT-TabGen \cite{chen2020logical}; and combining single-hop questions with sentences about the same entities to produce multi-hop questions via filling in masked tokens of designed templates. 
Reasoning graphs then define different types of reasoning chains for multi-hop QA using the operators as building blocks. Training QA pairs are generated by executing the reasoning graphs, which generate output texts. Finally, question filtration removes irrelevant and unnatural QA pairs to produce the final generated training set for multi-hop QA. The filtration is done by choosing the samples ranked as most fluent by GPT-2, and paraphrasing each generated question using BART.


    
\subsubsection{Sentiment Analysis (SA)} 
\citet{yu2021cross} applies data augmentation for aspect-based SA in the unsupervised domain adaptation setting, aiming to transform labeled datasets in a source domain to a new target domain. The main approach involves two steps. 
In the first step of domain generalization, domain-specific words and phrases in the labeled source data and unlabeled target data are identified and masked
in the inputs. Opinion words for the source domain and target-specific terms and opinion words are retrieved via sentiment lexicon and bootstrapping methods using relations in dependency trees. The target-specific terms in the unlabeled data will be masked to fine-tune BERT.
In the second step of {\it domain specification}, the source-specific terms in the source data are masked (thus producing domain-independent texts) and sent into the fine-tuned BERT to produce labeled sentences in the target domain. Here, some constraints based on dictionaries are necessary to ensure that the infilled words are terms or opinion words with the same sentiment polarity. The generated data can be used independently or combined with original source training data to train a SA model for the target domain. 

\citet{li2020conditional} use PLMs to generate synthetic data for aspect term extraction (cast as a sequence labeling problem). To fine-tune PLMs with the sequence-to-sequence framework for this purpose, the input includes a masked sentence from a training dataset and the corresponding label sequence while the output are the masked tokens in the input. The fine-tuned PLMs are then exploited to generate new possibilities for the masked tokens that can be injected into the masked input, using the original label sequence to obtain synthetic labeled data to train models.


\subsubsection{Fact Verification}
Fact verification aims to predict whether a given claim is supported, denied, or unresolved based on the given evidence. Automatically generated texts can be used to generate claim-evidence pairs for each label category. To this end, \citet{pan2021zero} employ a two-step approach to generate synthetic data for fact verification. In the first step of question generation, given the evidence and an answer, a BART model, fine-tuned on the SQuAD dataset using the similar input-output format, generates a question for that answer. Next, a question-to-claim model is employed to take the question and answer as inputs and generate a claim (also using a BART model fine-tuned on SQuAD). To produce $\langle$claim, evidence$\rangle$ pairs with the ``support'' relation, an entity is selected in the original evidence in the first step of the process. To produce a ``refute'' claim, the work replaces the original answer with another entity in the generation process. Finally, to create a ``not-enough-evidence'' claim, the paper expands the original evidence to include other paragraphs in the same document and produce claims for some entity in the extended paragraph. Experiments show competitive results when the augmented data is combined with few or even no human-labeled examples for model training.


\subsubsection{Document Classification}
A typical approach to generating synthetic data for text classification is to build a conditional generative model for each class by fine-tuning with labeled data from that class. While these models can be fine-tuned with the next word prediction objective with generative PLMs such as GPT-2, \citet{liu2020data}  use reinforcement learning to train generative models to augment text classification labeled data. The rewards for training are based on the similarity between the generated tokens and a salient lexicon of the target class computed via top frequency-based salient words, and the divergence between the conditional and unconditional models. \citet{liu2020data} demonstrate the effectiveness of using the automatically generated data in multiple text classification problems and datasets, including sentiment analysis and offense detection.




\subsection{Generating Auxiliary Data to Improve Different Aspects of NLP Models}


The following sections, again arranged by task, discuss ways of using PLM-generated text to aid in auxiliary tasks, helping developers or users understand model strengths and weaknesses or decision-making characteristics. 

\subsubsection{Explaining Models' Decisions}

Despite the impressive performance of deep learning models for various NLP tasks, a remaining challenge to widespread adoption is the lack of explanations for the models' decisions. This hinders the development and debugging process, as well as user trust. This is especially true for application domains such as healthcare, security, and online education. As such, a considerable number of approaches have been proposed for explaining deep learning models' behavior, including model-intrinsic \citep{ribeiro2016why,lundberg2017aunified,chen2018learning} and model-agnostic approaches \citep{park18multimodal,kim2018textual,ling2017program}. While model-intrinsic explanations expose internal model state (e.g. feature importance or attention scores), in model-agnostic (post-hoc) methods, explanations are generated via the model predictions without inspecting the internal state. Generative models are often applied for post-hoc explanations, aiming to obtain either counterexamples \citep{kim2016examples,wachter2018counterfactual,wu2021polyjuice} or natural language texts \citep{camburu2018esnli,kumar2020nile,chen2021kace} for explaining purposes. 

Generating {\it counterexamples} can shed light on the decision boundaries of the models (i.e. explaining when a model changes its decision), thus improving intepretability. To this end, the generated counterexamples should be close to the decision boundaries so that small modifications result in changing the model predictions. Traditionally, heuristic rules applied to the original inputs create likely counterexamples \citep{wachter2018counterfactual,ribeiro2018semantically,iyyer2018adversarial,li2021contextualized}. PLMs have been leveraged to generate more diverse examples for better evaluation \citep{madaan2021generate,wu2021polyjuice,ross2021explaining}. In particular, \citet{wu2021polyjuice} proposes a method based on GPT-2 to generate counterfactuals that are close to the original sentences and entail specific relationships with the original, facilitating label induction (e.g. negation, insertion, shuffle). Concretely, an input sentence is concatenated with a relation label (e.g. negation) and a template consisting of the special tokens \texttt{[BLANK]} to form the prompt for GPT-2 model. For instance, for the sentence ``\textit{It is great for kids}" and the relation label ``\texttt{negate}", the following prompt is constructed: ``\texttt{It is great for kids. [negation] It is [BLANK] great for [BLANK]. [SEP]}". Next, the GPT-2 model generates answers for the \texttt{[BLANK]} in the template (e.g. ``\texttt{not [ANSWER] children}'', separated by the special token \texttt{[ANSWER]}). To fine-tune the GPT-2 model, non-parallel datasets (e.g. CommonGen, Natural Questions and SQuAD) are automatically processed to find the relations between pairs of sentences and to construct the templates for each relation based on the obtained pairs. It is worth noting that the sentences generated by GPT-2 might have the same label as the original input sentence. In addition, \citet{wu2021polyjuice} show that the generated counterexamples can be helpful to improve the performance of the downstream models, e.g.~for natural language inference, duplicate question detection, and sentiment analysis.

Other research is informing the task of {\it natural language explanation generation}, where the goal is to expose the rationale behind the model decisions in automatically generated natural language text.  Any approach must critically require that the generated response is faithful to the model behavior. To this end, \citet{kumar2020nile} propose to first generate the explanations, and then employ the explanations to obtain the final model predictions. They use natural language inference as the task requiring explanations. Label-specific GPT-2 models are fine-tuned over concatenations of corresponding premises, hypotheses, and human-provided explanations, so that at inference, the model generates an explanation based on premise and hypothesis. Next, the explanations together with the premise and the hypothesis are consumed by an explanation processor model (e.g. RoBERTa) to select the most likely label. This process obtains a more faithful explanation for the  label choice, compared to traditional prediction-first approaches \cite{camburu2018esnli}. However, this approach does not provide explanations that reference non-selected labels. To address the question of why other labels are not chosen, \citet{chen2021kace} exploit counterexamples, deriving them from original samples with heuristic rules. The original samples and counterexamples are provided to GPT-2 to generate an explanation for the question ``\textit{Why A not B}''.

\subsubsection{Knowledge Extraction}

Generative PLMs are pre-trained on massive text corpora containing a large amount of information about entities and commonsense knowledge. As such, PLMs might directly be used to elicit knowledge required for downstream applications such as information extraction, sentiment analysis and question answering. To this end, it is important to properly prompt these models so their outputs contain the required information. Section~\ref{subsec:template-based_learning} describes the prompt design for knowledge extraction/probing tasks, and in particular, the ``Knowledge Probing'' subsection describes applications in details. Here we focus on the text generation aspect of knowledge extraction approaches. 

Prior works can be categorized into two sub-categories. The first category involves prompting PLMs with partial knowledge via a prompt and asking the models to complete the prompt. Specifically, pre-defined templates can be designed and filled with partial knowledge (e.g. the two entities involved in a relation) and the generative PLMs can predict the missing words in the templates (e.g. the relation type between the two entities.) The templates can be fixed \cite{goswami2020unsupervised} or they can be dynamically constructed by a pre-trained model \cite{shin-etal-2020-autoprompt} (further details are in Section  \ref{subsec:template-based_learning}). 
The second category instead proposes to prompt the PLMs with full knowledge and ask the models to generate a natural language text to describe that knowledge. This task is  known as Data-to-Text \cite{kukich1983design}, and the goal is to obtain a textual description of existing knowledge bases. The generated textual descriptions can be used by downstream applications such as knowledge probing \cite{petroni2019language} or QA \cite{agarwal2021knowledge}, among others. \citet{agarwal2021knowledge} introduce a model based on T5 to convert Wikidata knowledge graphs (with triples of relations between two entities) into textual data. The proposed approach consists of three stages. First,  create a large but noisy training dataset using distant supervision for relation extraction by aligning knowledge base (KB) triples to Wikipedia texts. Next, fine-tune T5 in stages, starting with the distantly supervised dataset for better coverage, then moving on to a small clean dataset for less hallucination. The model learns to generate descriptive sentences from KB triples. Last, build a filter for the generated texts based on semantic quality with respect to the KB triples by scoring the concatenation of input and output with BERT.

\subsubsection{Question Generation}

While PLMs can be directly used for generating answers for questions, they might be also helpful to support existing QA systems. Specifically, PLMs can be employed to provide clarification for downstream QA systems. The clarification can be realized in terms of question clarification when the question is ambiguous or it can be fulfilled by providing more context. For instance, in \citet{min2020answering} and \citet{min2020ambigqa}, multi-step question generation approaches are proposed for ambiguous QA in which the BART model is prompted with an ambiguous question and the top similar passages retrieved in a document to generate candidate answers. If multiple answers are generated, another BART model is employed to generate a disambiguation question for each answer. The newly generated questions are later used to extract other candidate answers. Finally, the generated answer-question pairs are ranked to select the top one for the ambiguous QA problem. \citet{min2020ambigqa} show that the process of generating auxiliary disambiguation questions could further help the models to encode the interactions between the original input question and the candidate answers. 

In another line of work, \citet{mao2021generation} seek to generate clarification texts for input questions to improve the retrieval quality in open-domain QA (answering factoid questions without a pre-specified domain). The most common approach for this problem involves a retriever-reader architecture \cite{chen2017reading}, which first retrieves a small subset of documents in the pool using the input question as the query and then analyzes the retrieved documents to extract (or generate) an answer. To generate augmented texts for the input question in the first retrieval component, \citet{mao2021generation} fine-tune BART to consume the input question and attempt to produce the answer and the sentence or title of the paragraph containing the answer. This method demonstrates superior performance for both retrieval and end-to-end QA performance.

In addition to clarification information, PLMs can also be used to paraphrase questions to support QA models. \citet{mass2020unsupervised} explore the problem of FAQ retrieval, retrieving the top QA pair given a user query. Based on the returned QA pairs $(q,a)$ from a retrieval system, this work proposes an unsupervised method to re-rank the pairs to improve the performance. One of the ranking scores is a matching score between the question $p$ in the pair $(q,a)$ with respect to the user question. A triple network is trained over the tuples $(p,q,q')$, where $q$ is a paraphrase of the question $p$ while $q'$ is randomly selected questions from other QA pairs. To this end, \citet{mass2020unsupervised} fine-tune GPT-2 over the concatenations of the corresponding answers and questions in the FAQ. The fine-tuned GPT-2 is then prompted with the answer $a$ to produce a paraphrase $q'$ for $q$ in the ranking network.


\subsubsection{Inference Rule Generation}

For some applications, it is important to understand the process by which the final predictions of the models are obtained. These intermediate inference rules provide are another form of model explanation and provide insights for improving model performance. 

\citet{paul2021coins} exploit GPT-2 to perform narrative story completion: given a few sentences of a story, the goal is to complete the story using sentences that logically follow the narrative in the given incomplete story. In an incremental generation method, each step seeks to generate a contextualized inference rule conditioned on the current incomplete story. To accomplish this, GPT-2 is fine-tuned on human annotation of story line inferences. Next, given the current story and generated inference rule, a new sentence for the story is generated (using another fine-tuned GPT-2 model). By interspersing the inference rules, the storyline generations should create a coherent story that follows logical connections and causal relationships between events.  

\citet{madaan2021could} employ T5 to generate inference graphs for defeasible inference \cite{rudinger2020thinking}. In this mode of reasoning, given a premise, a hypothesis may be weakened or overturned in light of new evidence. As training inference graphs for this problem requires a large amount of human-annotated inference graphs, they propose to exploit reasoning graphs in related tasks to fine-tune T5. In particular, this work leverages the influence graphs in the WIQA dataset that includes a set of procedural passages, each accompanied by a human-curated influence graph. The influence graphs are linearized to fit into the seq2seq framework for fine-tuning T5 and producing inference graphs for defeasible inference afterward. It has been shown that the generated inference graphs can improve human accuracy on defeasible inference (which is originally challenging for humans).

\section{Discussion}
\label{sec:discussion}

\paragraph{Mix of paradigms or PLMs.} The three paradigms presented in this paper are by no means mutually exclusive. Instead, it is not rare to see approaches that use two or three paradigms together: fine-tuning techniques are often used as part of prompt-based methods; NLP-as-text-generation approaches often use carefully crafted templates (prompts); and prompt-based learning often leverages the text generation capabilities of PLMs to generate words, phrases, or sentences.  

A representative example is  \citet{khashabi-etal-2020-unifiedqa}, which combined three paradigms:  appropriate prompts from the context and questions help to formulate several QA tasks into a unified text generation problem with seq2seq-based pre-trained models such as T5, with model fine-tuning  to improve performance in several QA tasks.

As independently trained models, PLMs are also by no means mutually exclusive. For example, ACE~\cite{wang2021automated} shows that combining multiple PLMs (e.g ELMo, BERT, mBERT, XLM-R) yields further improvements over using a single PLM for a range of NLP tasks. Investigation of the complementarity of different PLMs is a future research direction. 

From another perspective, the design of the training for MLMs has been driven by the results on the fine-tuning paradigm, but it is not clear whether an exploration of different training objectives could lead to PLMs that are more effective when used with prompting or generation to solve NLP tasks. 

\paragraph{How much unlabeled data is needed?}
While PLMs are usually trained on billions of words, some works have investigated what can be learned with less pre-training data. \citet{zhang-etal-2021-need}, experimenting on RoBERTa models trained on 1M, 10M, 100M and 1B words \citep[MiniBERTas]{warstadt-etal-2020-learning}, showed that 10M to 100M words are sufficient to acquire many syntactic and semantic features. \citet {HSFR21} presented BabyBERTa, a RoBERTa-based model trained on language acquisition data that acquires grammatical knowledge comparable to that of pre-trained RoBERTa-base -- and does so with approximately 15x fewer parameters and 6,000x fewer words. On the other hand, \citet{zhang-etal-2021-need}, using the pre-train then fine-tune paradigm for NLU tasks, found that millions of words are not sufficient for key NLU skills, which instead may require billions of words and continue improvements with additional  pre-training data.

\paragraph{How much labeled data is still needed?}
While \citet{lescao2021many} present experiments to quantify the impact of prompts, there has been little work in designing rigorous experiments to study how many labeled examples are required by PLMs to achieve various levels of performance for a range of NLP tasks, and using each of the three paradigms outlined in this survey. Such studies will provide a better understanding of the pros and cons of each formulation, including cost-benefit analyses weighing the impact of more labeled data, helping developers design NLP systems that achieve the desired goal while minimizing human labeling effort.

\paragraph{Can we reduce the amount and cost of computation?}
The development of deep learning in general and the use of PLMs in particular have dramatically increased the amount of computation used in NLP, leading to a high environmental footprint. \citet{schwartz2020green} argue for Green AI, suggesting that we should consider efficiency, measured by the number of floating-point operations used to generate a result, as a main evaluation criterion, together with accuracy. Green AI also aims to reduce the financial cost of the computation. In line with this approach, \citet{izsak2021train} propose software optimization and design choices for pre-training BERT in 24 hours using a single low-end deep learning server.

\paragraph{Do PLMs excel at semantic understanding or memorization?}
Another interesting avenue to explore is separating extraction or text understanding from memorization. To what extent can PLMs memorize facts and extract an answer from a passage provided (understanding a text), for knowledge-intensive tasks such as Questions Answering (QA) and Information Retrieval (IR)? This is motivated by the observation by ~\citet{wang-etal-2021-generative} that PLMs are terrible at remembering training facts with high precision and that it is also challenging for them to answer closed-book questions even if relevant knowledge is retained.

\paragraph{Is explicit linguistic information needed?}
A related debate is whether a symbolic annotation covering syntax or semantics should be integrated to improve the performance of a PLM-based system, or whether this information is already present in the model. Below we list some successes in leveraging syntax or semantics, though there is no definite answer yet. In terms of syntax, \citet{xu-etal-2021-syntax} utilize automatically produced syntax in both the pre-training and fine-tuning stages, and show improved performance on several benchmark datasets. \citet{nguyen2020treestructured} and \citet{sachan-etal-2021-syntax} inject syntax only in the fine-tuning stage.  Regarding semantics, \citet{zhang2020semantics} incorporate Semantic Role Labeling predictions into the pre-training procedure of BERT, improving the performance on textual entailment and QA tasks. \citet{wu2021infusing} integrate semantic information into the task-specific fine-tuning stage, focusing on the DELPHIN dependencies formalism or ``DM" \cite{ivanova-etal-2012-contrastive}. Experimenting on RoBERTa, they obtained improvements on the GLUE benchmark.
Syntax and semantics can also be jointly integrated, as in \citet{zhou-etal-2020-limit}, where multi-task learning was used to combine BERT pre-training with both semantic and syntactic parsing tasks, improving the performance on the GLUE benchmark.

\paragraph{Can we integrate implicit semantic information using QA?} 

Instead of enriching PLMs with symbolic annotations, a possible alternative for a supervision signal is QA data, as it is easier to answer questions relative to a sentence than to annotate linguistic phenomena in it \citep{roth2017incidental,he-etal-2020-quase}.
In the s-QuASE PLM presented in \citet{he-etal-2020-quase}, further pre-training of BERT on QA datasets is done while restricting the interaction between the question and context inputs.
s-QuASE is particularly useful in single-sentence tasks such as Semantic Role Labeling and NER.
A similar direction was pursued by \citet{jia2021question} who leveraged question generation and knowledge distillation to build a QA-based pre-training objective.

\paragraph{Do PLMs need meaningful prompts?} The success of prompts in zero- and few-shot learning has been attributed to the prompts serving as instructions that allow the PLM to learn with fewer examples, much the way humans would \cite{mishra2021crosstask,schick-schutze-2021-exploiting,NEURIPS2020_1457c0d6}. In fact, the excellent results may instead be attributable to the mere exploitation of patterns in the training data of PLMs, and not to PLMs' perceived ability to interpret and follow meaningful instructions. ~\citet{webson2021promptbased} show, for instance, that irrelevant templates match the performance of meaningful ones in few-shot entailment experiments, adding that some of the templates discovered by automatic generation of discrete prompts are also unnatural \cite{shin-etal-2020-autoprompt}. In this sense, the results of continuous prompts also show that PLMs do not need meaningful instructions for improving few-shot performance.

\paragraph{Theoretical and empirical analysis}
The theoretical understanding of the paradigms presented in this survey is preliminary. Apart from the issues mentioned above, there is a lack of understanding of what actually makes these paradigms so successful, and whether their success can be generalized across models and languages. For instance, prompts may be PLM-dependent, or they may be transferable across models as indicated in \cite{perez2021true}. There is very little work on studying the generalization of prompting and generation across languages, in the way that transfer learning has been applied to learning in one language and testing in another \cite{conneau2020unsupervised}.  

\section{Conclusion}
\label{sec:conclusion}

In this paper, we present a survey of the three trending paradigms that use pre-trained language models for NLP. We describe each of them in depth, and summarize prior works whose applications have shown promise. In addition, we describe the use of pre-trained language models to automatically generate data that is used to improve performance in NLP tasks. We hope this survey will provide readers with key fundamental concepts and a comprehensive view of the paradigm shift. 

\section*{Acknowledgments}

This research is based upon work supported in part by the 
Office of the Director of National Intelligence (ODNI),
Intelligence Advanced Research Projects Activity (IARPA),
via Contract No. 2019-19051600006 under the IARPA BETTER program and by Contracts FA8750-
19-2-0201 and FA8750-19-2-1004 with the US
Defense Advanced Research Projects Agency
(DARPA). Approved for Public Release, Distribution Unlimited.  The views and conclusions 
contained herein are those of the authors and should not be 
interpreted as necessarily representing the official policies, 
either expressed or implied, of ODNI, IARPA, the Department of Defense or the U.S.
Government. The U.S. Government is authorized to reproduce 
and distribute reprints for governmental purposes not 
withstanding any copyright annotation therein.

We would like to thank Paul Cummer for his insightful comments on this work.

\bibliographystyle{acl_natbib}
\bibliography{anthology,acl2021}

\newpage

\appendix
\section{PLMs for specialized domains or languages} \label{app:plms}

Table~\ref{tab:domain_plm} shows PLMs for special domains. Table~\ref{tab:other_lang_plm} presents PLMs pre-trained on different languages.

\begin{table*}[h!]
    \centering
    \scalebox{0.75}{
\begin{tabular}{|p{0.4\linewidth}|p{0.25\linewidth}|p{0.6\linewidth}|}
\hline
\textbf{Model} & \textbf{Domain} & \textbf{Training Sources} \\
\hline
\textsc{SciBERT}~\citep{beltagy2019scibert} & Science &  Scientific articles in computer science and biomedicine  \\
\hline
\textsc{BioBERT}~\citep{jinhyuk2019biobert} & Biomedical & Biomedical publications  (abstracts and full-text articles) \\
\hline
\textsc{ClinicalBERT}~\citep{huang2020clinicalbert}, \citet{alsentzer-etal-2019-publicly} & Clinical & Clinical notes \\
\hline
\textsc{LegalBERT}~\citep{chalkidis-etal-2020-legal} & Legal & Legal documents (e.g. contracts) \\
\hline
\textsc{CodeBERT}~\citep{feng-etal-2020-codebert}, \textsc{Codex}~\citep{chen2021evaluating} & Source code & GitHub repositories \\
\hline
BERTweet~\citep{nguyen-etal-2020-bertweet}, AlBERTo~\citep[for Italian;][]{Polignano2019AlBERToIB} & Twitter & Tweets \\
\hline
BabyBERTa~\citep{HSFR21} & Child-directed speech &  Child-directed speech transcriptions\\
\hline 
\end{tabular}}
    \caption{PLMs pre-trained on specific domains.}
    \label{tab:domain_plm}
\end{table*}

\begin{table*}[]
    \centering
    \scalebox{0.75}{
\begin{tabular}{|l|l|}
\hline
\textbf{Language} & \textbf{Model}\\
\hline
Arabic & Arabic-BERT~\cite{safaya-etal-2020-kuisail} \\
\hline 
Basque & BERTeus~\cite{agerri-etal-2020-give} \\
\hline 
Chinese & MacBERT~\cite{cui-etal-2020-revisiting}\\
\hline
Dutch & BERTje~\cite{devries2019bertje}, RobBERT~\cite{delobelle-etal-2020-robbert} \\
\hline
Farsi & ParsBERT~\cite{farahani2021parsbert} \\
\hline
Finnish & FinBERT~\cite{virtanen2019multilingual} \\
\hline
French & CamemBERT~\cite{martin-etal-2020-camembert}, FlauBERT~\cite{le-etal-2020-flaubert-unsupervised} \\
\hline
German & GBERT and GELECTRA~\cite{chan-etal-2020-germans} \\
\hline
Hebrew & HeBERT~\cite{chriqui2021hebert}, AlphaBERT~\cite{seker2021alephberta} \\
\hline
Italian & GilBERTo ~\cite{ravasio2020gilberto}, UmBERTo ~\cite{parisi2020umberto}\\
\hline
Japanese & Japanese BERT \cite{inui2021pretrained}\\
\hline
Portuguese & BERTimbau ~\cite{souza2020bertimbau}\\
\hline
Russian & RuBERT~\cite{kuratov2019adaptation}\\
\hline
Spanish & BETO~\cite{canete2020spanish} \\
\hline
Turkish & BERTurk ~\cite{schweter2020berturk}\\
\hline
\end{tabular}}
    \caption{PLMs pre-trained on different languages.}
    \label{tab:other_lang_plm}
\end{table*}

\section{Pre-train then fine-tune approaches} \label{app:pre-train-finetune-approaches}

Table~\ref{tab:prior_work_by_finetuning_strategy} shows a summary of prior work organized by the strategies in the first paradigm, ``pretrain then fine-tune''. It is worth noting that ``Contextual embeddings'' does not involve fine-tuning, but we included here because it is architecturally similar to other strategies, except that the PLM's weights are frozen, that is they are not fine-tuned for the specific tasks.

\begin{table*}[]
    \centering
    \scalebox{0.65}{
\begin{tabular}{|p{0.55\linewidth}|p{0.65\linewidth}|p{0.25\linewidth}|}
\hline
\textbf{Task} & \textbf{Work} & \textbf{PLM} \\
\hline
\multicolumn{3}{|c|}{(1) \textbf{Contextual Embeddings}} \\
\hline
Word Sense disambiguation/induction &  ~\citet{hadiwinoto-etal-2019-improved}, ~\citet{amrami2019better} & BERT \\
Coreference resolution & ~\cite{lee-etal-2018-higher} & ElMO \\
Constituency parsing & ~\cite{zhang2020fast} & BERT \\
 & ~\citet{yang2020strongly} & XLNet, BERT \\
& ~\citet{zhou-zhao-2019-head} & ELMo, BERT  \\
Constituency parsing, Dependency parsing & ~\cite{mrini-etal-2020-rethinking} & XLNet, BERT \\
Dependency parsing & ~\citet{wang-tu-2020-second} & BERT \\
& ~\citet{schuster-etal-2019-cross} & ElMo \\
Semantic role labeling & ~\cite{he-etal-2018-jointly} & ElMO \\
AMR parsing & ~\citet{cai-lam-2020-amr}, ~\citet{xu-etal-2020-improving} & BERT \\
UCCA parsing & ~\cite{jiang-etal-2019-hlt} & BERT, mBERT \\
Commonsense reasoning & ATOMIC~\cite{sap2019atomic} & ELMo \\
Machine Translation & ~\citet{zhu2020incorporating} & BERT \\
Text summarization & ~\cite{zhang-etal-2019-pretraining} & BERT \\
\hline
\multicolumn{3}{|c|}{(2) \textbf{Fine-tuning the PLM}} \\
\hline
Text classification & ~\cite{yang2019xlnet} & XLNet \\
& ~\cite{sun2020finetune} & BERT \\
& ~\cite{peters2018deep} & ElMO \\
Semantic textual similarity & ~\cite{yang2019xlnet} & XLNet \\
NER & ELMo~\cite{peters2018deep} & ELMo \\
NER, QA, Textual Entailement (TE) & ~\citet{devlin-etal-2019-bert} & BERT \\
TE & ~\cite{liu2019roberta} & RoBERTa \\
Entity Linking &  ~\citet{broscheit-2019-investigating}, ~\citet{ling2020learning}, ~\cite{wu2020scalable} & BERT \\
Relation extraction & ~\cite{baldini-soares-etal-2019-matching, wu2019enriching, shi2019simple} & BERT \\
Intent Detection and Slot Filling & ~\cite{chen2019bert} & BERT  \\
& XLNet~\cite{yang2019xlnet} & XLNet \\
Text generation & ~\cite{kale-rastogi-2020-text} & T5 \\
Coreference resolution & ~\cite{joshi2019bert} &  BERT \\
& ~\cite{yu2020paired} & RoBERTa \\
Text simplification & ~\cite{martin2021muss} & BART/mBART \\
& ~\cite{raffel2020exploring} & T5 \\
Dialogue & ~\cite{hosseiniasl2020simple} & GPT-2 \\
Semantic role labeling & ~\cite{shi2019simple} & BERT \\
Text summarization & ~\cite{liu2019text} & BERT \\
& ~\cite{lewis-etal-2020-bart} & BART \\
Commonsense reasoning & COMET~\cite{bosselut2019comet} & GPT \\ 
& ATOMIC2020~\cite{hwang2020cometatomic} & GPT2 \\
Machine Translation & ~\cite{liu2020multilingual} & mBART \\
& ~\cite{lample2019crosslingual} & XLM \\
\hline
\multicolumn{3}{|c|}{(3) \textbf{Fine-tuning Customized Models}} \\
\hline
NER, POS tagging, dependency parsing, aspect extraction &  ACE~\cite{wang2021automated} & ELMo, (m)BERT, XLM-R \\
Semantic parsing &  ~\cite{che-etal-2019-hit} & BERT \\
Temporal relation extraction & ~\cite{ross-etal-2020-exploring} & BERT \\
Text simplification & ~\cite{omelianchuk-etal-2021-text} & RoBERTa \\
Text simplification, summarization & ~\cite{malmi-etal-2019-encode} & BERT \\
Coreference resolution & CorefQA~\cite{wu-etal-2020-corefqa} & SpanBERT \\
Machine Translation & ~\cite{weng2020acquiring} & BERT, GPT \\
& ~\cite{ma2020xlmt} & XLM-R \\
CCG parsing & Tree-structured supertagger~\cite{TACL2445} & RoBERTa-base \\
\hline
\multicolumn{3}{|c|}{(4) \textbf{Efficient Fine-tuning Approaches}} \\
\hline
& BitFit~\cite{zaken2021bitfit} & \\
& Adapter-Transformer~\cite{pfeiffer-etal-2020-adapterhub} & \\
POS tagging, dependency parsing &  Trankit~\cite{nguyen2021trankit} & XLM-R \\
\hline
\end{tabular}}
    \caption{A summary of prior work organized by the strategies in the first paradigm ``pretrain then fine-tune''.}
    \label{tab:prior_work_by_finetuning_strategy}
\end{table*}

\end{document}